%% file: ms.tex
\documentclass[letterpaper,twocolumn,10pt]{article}
\usepackage{usenix}

\usepackage{tikz}
\usepackage{amsmath}

\usepackage[normalem]{ulem}
\usepackage{xspace}
\usepackage{comment}
\usepackage{ifthen}
\usepackage{caption}
\usepackage{subfig}
\usepackage{booktabs}

\usepackage{listings}

\lstdefinelanguage{RAF}
{
  morekeywords={
    fn,
    let,
    Tensor,
  },
  sensitive=true, 
  morecomment=[l]{//}, 
  morecomment=[s]{/*}{*/}, 
  morestring=[b]" 
}

\definecolor{codegreen}{rgb}{0,0.6,0}
\definecolor{codegray}{rgb}{0.5,0.5,0.5}
\definecolor{codepurple}{rgb}{0.58,0,0.82}
\definecolor{backcolor}{rgb}{0.95,0.95,0.92}
\definecolor{eclipseBlue}{RGB}{42,0.0,255}
\definecolor{eclipsePurple}{RGB}{127,0,85}

\lstdefinestyle{mystyle}{
    language={RAF},
    commentstyle=\color{codegreen},
    keywordstyle=\color{magenta},
    numberstyle=\tiny\color{black},
    stringstyle=\color{codepurple},
    basicstyle=\ttfamily\footnotesize,
    emph={shape_of, alloc_storage, alloc_tensor, invoke_mut, invoke_shape_func},
    emphstyle=\color{eclipseBlue},
    breakatwhitespace=false,
    breaklines=true,
    captionpos=b,
    keepspaces=true,
    numbers=left,
    numbersep=4pt,
    showspaces=false,
    showstringspaces=false,
    showtabs=false,
    tabsize=2
}
\newcommand{\system}[0]{RAF\xspace}
\newcommand{\showcomments}{yes}
\newcommand\yida[1]{
    \ifthenelse{\equal{\showcomments}{yes}}{\textcolor{red}{[yida: #1]}}{\ignorespaces}
}
\newcommand\cody[1]{
    \ifthenelse{\equal{\showcomments}{yes}}{\textcolor{blue}{[cody: #1]}}{\ignorespaces}
}
\newcommand\haichen[1]{
    \ifthenelse{\equal{\showcomments}{yes}}{\textcolor{magenta}{[haichen: #1]}}{\ignorespaces}
}
\newcommand\zhen[1]{
    \ifthenelse{\equal{\showcomments}{yes}}{\textcolor{violet}{[Zhen: #1]}}{\ignorespaces}
}
\newcommand\todo[1]{
    \ifthenelse{\equal{\showcomments}{yes}}{\textcolor{red}{[TODO: #1]}}{\ignorespaces}
}
\newcommand\jie[1]{
    \ifthenelse{\equal{\showcomments}{yes}}{\textcolor{orange}{[jie: #1]}}{\ignorespaces}
}

\begin{document}

\date{}

\title{\Large \bf \system: Holistic Compilation for Deep Learning Model Training}

\author{
{\rm Cody Hao Yu}\\
Amazon Web Services, Inc
\and
{\rm Haozheng Fan}\\
Amazon Web Services, Inc
\and
{\rm Guangtai Huang}\\
Amazon Web Services, Inc
\and
{\rm Zhen Jia}\\
Amazon Web Services, Inc
\and
{\rm Yizhi Liu}\\
Amazon Web Services, Inc
\and
{\rm Jie Wang}\\
Amazon Web Services, Inc
\and
{\rm Zach Zheng}\\
Amazon Web Services, Inc
\and
{\rm Yuan Zhou}\\
Amazon Web Services, Inc
\and
{\rm Haichen Shen}\\
Scroll Foundation
\and
{\rm Junru Shao}\\
OctoML, Inc
\and
{\rm Mu Li}\\
Amazon Web Services, Inc
\and
{\rm Yida Wang}\\
Amazon Web Services, Inc
} 

\maketitle

\begin{abstract}
As deep learning is pervasive in modern applications, many deep learning frameworks are presented for deep learning practitioners to develop and train DNN models rapidly.
Meanwhile, as training large deep learning models becomes a trend in recent years, the training throughput and memory footprint are getting crucial.
Accordingly, optimizing training workloads with compiler optimizations is inevitable and getting more and more attentions.
However, existing deep learning compilers (DLCs) mainly target inference and do not incorporate holistic optimizations, such as automatic differentiation and automatic mixed precision, in training workloads.

In this paper, we present \system, a deep learning compiler for training.
Unlike existing DLCs, \system accepts a forward model and in-house generates a training graph.
Accordingly, \system is able to systematically consolidate graph optimizations for performance, memory and distributed training.
In addition, to catch up to the state-of-the-art performance with hand-crafted kernel libraries as well as tensor compilers, \system proposes an operator dialect mechanism to seamlessly integrate all possible kernel implementations.
We demonstrate that by in-house training graph generation and operator dialect mechanism, we are able to perform holistic optimizations and achieve either better training throughput or larger batch size against PyTorch~\cite{paszke2019pytorch} (eager and torchscript mode), XLA~\cite{xla}, and DeepSpeed~\cite{he2016deep} for popular transformer models on GPUs.
\end{abstract}

\input{Sections/1.intro}

\input{Sections/2.background}

\input{Sections/3.system}
\input{Sections/4.impl}

\input{Sections/5.evaluation}
\input{Sections/6.related}

\input{Sections/7.conclusion}

\bibliographystyle{plain}
\bibliography{references}

\end{document}

%% file: Sections/1.intro.tex
\section{Introduction}
\label{sec:intro}
In recent years, deep learning is pervasive in modern applications, ranging from computer vision~\cite{he2016deep, krizhevsky2012alexnet,  szegedy2015going, szegedy2016rethinking}, nature language processing~\cite{devlin2018bert, radford2019language, vaswani2017attention}, to speech recognition~\cite{chan2021speechstew, chung2021w2v, zhang2020pushing}.
As deep learning models along with their training datasets are getting larger and larger, the efficiency of training deep learning models becomes more and more critical.
Modern deep learning practitioners normally rely on the deep learning frameworks, such as TensorFlow~\cite{abadi2016tensorflow} and PyTorch~\cite{paszke2019pytorch}, to describe and train models.
However, deep learning frameworks are not designed for efficient model training in the first place.
Instead, they aim for friendly interactive experience to facilitate model design.
When it comes to performance, frameworks normally invoke hand-crafted kernel libraries such as cuDNN~\cite{chetlur2014cudnn} to execute the computationally-intensive operators.
This is constrained by whatever the kernel libraries can provide, and misses the global optimization opportunities between operators (e.g., operator fusion and decomposition).

Moreover, the design of modern deep learning frameworks also limits the developments of distributed training systems~\cite{li2020pytorchdist,rasley2020deepspeed,zhang2022mics}, which are usually built on the top of these frameworks.
Taking DeepSpeed~\cite{rasley2020deepspeed}, a state-of-the-art (SOTA) distributed system built on top of PyTorch, as an example: although DeepSpeed implements a distribution engine with ZeRO~\cite{rajbhandari2020zero} memory optimization technique to enable gigantic model training, its per-device execution engine is native PyTorch runtime without graph optimizations.
As a result, DeepSpeed cannot globally optimize the model execution by well hiding the latency of inter-device communication~\footnote{PyTorch features ``hooks'' that allow developers to inject callback functions before and after each tensor or module execution, so it is possible to prefetch tensors using hooks to overlap communication latency. However, it is challenging to derive an optimal prefetch plan for various model architectures without analyzing an entire model graph.} during distributed training.

Consequently, adopting deep learning compilers (DLCs) to be the backend engine of deep learning frameworks become a promising solution to efficient model training, as they are capable of systematically optimizing the entire model graph and generating kernel code for the hardware platform.
For example, PyTorch incorporates torchscript as its compiler backend for model inference, which recently was extended to support training workloads as well and enhanced by nvFuser~\cite{nvfuser}.
However, torchscript itself does not cover distributed computation, but relies on other components in PyTorch to do so.
XLA~\cite{xla} is the most notable deep learning compiler that tackles model training, and some deep learning frameworks have adopted XLA as their compilation engine.
For instance, JAX~\cite{jax2018github} uses XLA to just-in-time (JIT) compile pure-and-statically-composed (PSC) subroutines; PyTorch also supports XLA of its traced model~\cite{suhan2021lazytensor} to enable model training with compilation.
However, most existing compilers like XLA were not designed in a \emph{holistic} way to facilitate full stack optimization.
For instance, XLA relies on the framework for part of graph-level manipulations such as auto-differentiation and automatic mixed precision (AMP).
Furthermore, XLA does not guarantee that the best available kernel implementations can be used.

We argue that the best deep learning model training performance can be achieved by taking full control of the entire software stack from graph-level to operator-level. 
This paper proposes \system as a compiler-based system that provides compilation for deep learning model training with holistic optimizations.
We use \system to demonstrate the following points.

\noindent\textbf{\#1: Holistic optimization.} 
\system traces vanilla deep learning models~\footnote{A vanilla model is a user-written model without any framework specific manipulations, such as auto-diff and automatic mixed precision.} from a framework such as PyTorch~\cite{paszke2019pytorch} to generate required training graphs and compile them all the way to the executables, including graph manipulation and optimization, operator-level kernel code generation, and distributed parallelism implementation.
We will illustrate in \autoref{sec:eval} that holistic optimization leads to a better performance.


\noindent\textbf{\#2: Three-phase graph optimizations.} 
\system for the first time puts together three types of graph-level optimization into one compiler stack, including graph generation (e.g., backward graph by automatic differentiation), expression optimizations (e.g., constant folding), and execution order optimizations (e.g., memory planning).
Accordingly, \system systematically abstracts the graph-level optimizations to three phases, and chooses the most suitable intermediate representation (IR) for each phase to ease the developments. 
While existing DLCs either work on dataflow~\cite{chen2018tvm} or A-normal form (ANF)~\cite{xla} IR for all optimization passes for simplicity, we will illustrate that 1) adopting the suitable IR form in each optimization phase could improve the development efficiency; and 2) it is possible to preserve the semantic when converting IRs between dataflow and ANF forms.

\noindent\textbf{\#3: Extensible backend for kernel libraries and tensor compilation.}
To make use of hand-crafted kernels from different backends, \system introduces \textit{an operator dialect mechanism} to dispatch each operator, which could be a single operator or a fused computation subgraph, to either high-performance kernel libraries or a tensor program compiler.
We demonstrate that by intelligently dispatching to high quality hand-crafted kernels while evolving the tensor compiler, \system could catch up with the latest SOTA performance at all time.


We summarize contribution of this paper as follows:
\begin{itemize}
    \item We design and implement \system training compiler that performs holistic optimizations to transfer a vanilla model all the way to an executable.
    \item We conduct comprehensive graph-level optimizations in three phases -- graph generation, expression optimization, and execution order optimization, and adopt the most suitable IRs in each phase for better development efficiency.
    \item We introduce an operator dialect mechanism to enable intelligent operator dispatching, including third-party kernel libraries and a tensor compiler, for multiple platforms.
    \item We evaluate \system with popular transformer models and show that \system either outperforms the SOTA performance or achieves larger batch sizes on a single and multiple GPUs, respectively.
\end{itemize}

The source code of \system is open source at \url{https://github.com/awslabs/raf}.

%% file: Sections/2.background.tex
\section{Background}
\label{sec:bg}

In this section, we first analyze the difference between training and inference workloads in \autoref{sec:bg_train}, followed by an introduction to deep learning compilers (DLCs) with the challenges of supporting training workloads in \autoref{sec:bg_dlc}.

\subsection{Deep Learning Training Workloads}
\label{sec:bg_train}

\input{Figures/fig-train-workload}

This paper focuses compilation technologies to enable better throughput and memory footprint of deep learning model training.
We illustrate the difference between inference and training workloads along with  \autoref{fig:train_workload}.
An inference workload are composed of a forward graph defined in the deep learning models, as well as trained parameters.
Meanwhile, a training workload is composed of 1) a forward graph (\textbf{A} and \textbf{B}), 2) loss computation, 3) a backward graph (\textbf{C} and \textbf{D}), 4) an optimizer (\textbf{E} and \textbf{F}), and 5) learnable parameters being trained.
These differences make training workloads more challenging to be optimized by a deep learning compiler.

First, in inference workloads, since parameter values are trained and frozen, they are treated as constants. Thus, the performance overhead of related optimizations such as data layout (e.g., row-major or column-major) and data type (e.g., full precision, half precision or quantization) transformation can be eliminated via constant folding.
This, however, is not a case for parameters being trained in training workloads.
Second, unlike inference workloads that usually require single or small batches
and focus on latency optimization, training workloads focus on reducing convergence time~\cite{mattson2020mlperf}, which could be achieved by large batch sizes and high throughputs.
Consequently, optimizing memory footprint~\cite{chen2016training,gruslys2016memory,kirisame2020dynamic,kumar2019efficient} to support larger batch sizes, as well as optimizing distributed mechanism~\cite{rasley2020deepspeed,zhang2022mics,zheng2022alpa} to achieve a higher throughput, are the key for training workloads.


\subsection{Deep Learning Compiler}
\label{sec:bg_dlc}
Unlike deep learning frameworks that directly map operators to hand-crafted kernel libraries for execution, deep learning compilers (DLCs), such as Apache TVM~\cite{chen2018tvm} and XLA~\cite{xla}, serve as a backend engine that converts the model graph from deep learning frameworks to a particular intermediate representation (IR), applies a series of graph- and operator-level optimizations, and generates an executable.
However, it is challenging for DLCs to achieve the optimal performance or to be adopted in training workloads:


\noindent\textbf{Challenge~1: Optimization scope.} In addition to the forward inference, training graph also includes backward propagation and weight updating. Existing DLCs optimize the complete training graph generated by deep learning frameworks.
However, as we will illustrate in \autoref{sec:impl_autodiff}, framework-generated training graph may not be friendly to the DLC, which results in sub-optimal performance.

\noindent\textbf{Challenge~2: Intermediate tensors.} Unlike inference graphs that are mostly a straight dataflow so intermediate tensors can be released immediately, in training workloads, many intermediate tensors generated by forward inference have to be materialized and preserved for gradient calculations in the backward propagation.
For example, the output tensors of node \textbf{A} and \textbf{B} in \autoref{fig:train_workload} can be freed soon in inference, but they have to be alive until node \textbf{C} and \textbf{D} in training.
These long-life intermediate tensors introduce two challenges: 1) \textit{Memory capacity and optimization} become crucial. 2) The operators that produce and consume the tensor in the forward graph (e.g., node \textbf{A} and \textbf{B}) cannot be fused anymore, significantly \textit{reducing fusion opportunities}.

\noindent\textbf{Challenge~3: Distribution.} While most inference workloads still target a single device, modern DLCs that target inference workloads lack the distribution support and cannot be used for training workloads, which may require to be scaled out due to the size of deep learning model as well as the training datasets.

%% file: Figures/fig-train-workload.tex
\begin{figure}[!tb]
	\centering
	\includegraphics[width=0.8\linewidth]{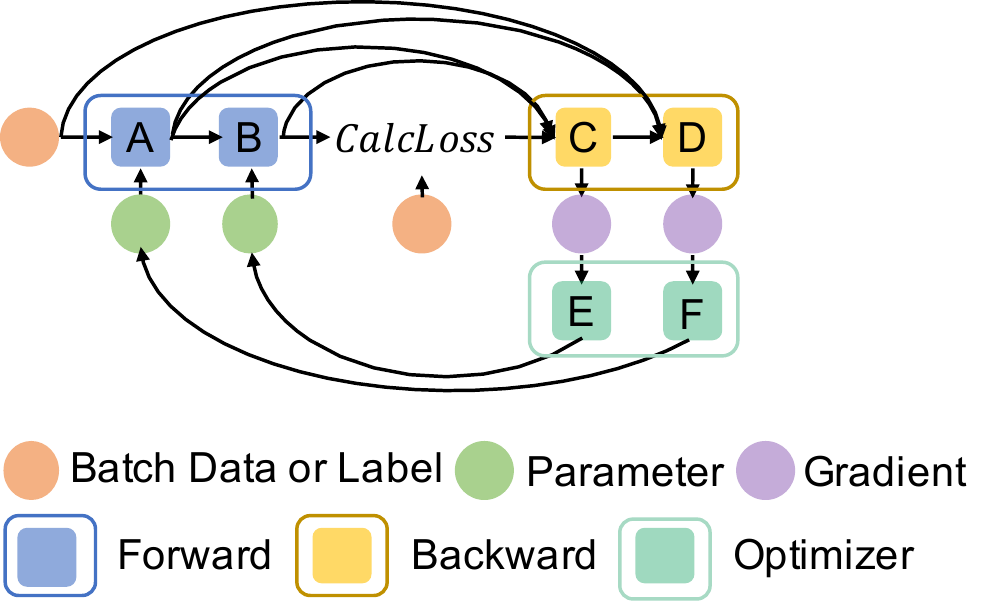}
	\caption{Deep learning training workload.}
	\label{fig:train_workload}
\end{figure}

%% file: Sections/3.system.tex
\section{System Design}
\label{sec:impl}

Existing deep learning compilers (DLCs) focus on optimizing a given IR without changing its semantics to limit the problem scope, so that they include only expression and execution order optimizations.
Accordingly, these DLCs can choose one IR format for all compiler passes to simplify the framework design.
For example, all compiler passes of XLA~\cite{xla} work on ANF IR.
Meanwhile, Apache TVM~\cite{chen2018tvm} lets almost all its compiler passes work on dataflow IR\footnote{Although TVM also converts dataflow IR to ANF, it is only for virtual machine execution.}, and lack of execution order optimization.

On the other hand, to include holistic optimization, \system accepts the IR from a vanilla forward (i.e., inference) model, and generates user-specified graphs (e.g., backward and automatic mixed precision (AMP)~\cite{amp} graphs).
Consequently, to deal with the new introduced complexity, we organize compiler passes in \system to three phases -- graph generation, expression optimization, and execution order optimization.
Figure~\ref{fig:overview} presents an overview of the \system compilation flow.
The first phase applies user-specified model manipulations to the graph in ANF IR. For instance, it may append backward propagation generated by auto-differentiation (\autoref{sec:impl_autodiff}), optimizer (e.g., SGD~\cite{bottou2010large} and Adam~\cite{kingma2014adam}), perform auto-casting to enable automatic mixed precision (AMP)~\cite{amp} training (\autoref{sec:impl_autocast}), or auto-parallelism (\autoref{sec:impl_dist}) to enable ZeRO~\cite{rajbhandari2020zero} data parallelism.
Since expressions in the ANF IR are bound to a variable using \texttt{let}, and all let-statements form a sequence that implies the execution order.
Therefore, it is easy to encode extra information in the IR associated with variables, such as the chain rule information in the automatic differentiation and parameter aliasing for in-place updating.


\input{Figures/fig-overview}

Next, in the optimization phase, we first convert the IR from ANF to dataflow, and apply a series of IR transformation passes to optimize the expression.
This is because the enforcement of expression order in ANF makes it harder for developers to implement transformation passes that focus on optimizing expressions but not execution order, such as operator fusion and expression simplification.
In contrast, dataflow IR forms a model to be a dataflow graph fashion, and expressions are embedded as arguments in the subsequent expressions.
As a result, it is straightforward to implement pattern- and rule-based expression transformations in dataflow IR.

Finally, in the phase of execution order optimization, we again convert the IR back to ANF and apply a series of passes. For instance, distribution passes (\autoref{sec:impl_dist}), including collective communication operator optimization and computation-communication overlapping, are also used to empower gigantic model training.
Meanwhile, memory optimization passes, including memory-footprint-aware execution order scheduling (\autoref{sec:impl_op_sche}) and automatic rematerialization (\autoref{sec:impl_remat}), are applied to reduce peak memory consumption to support larger batch sizes.
Note that since dataflow IR does not bound expressions with variables, we encode extra information in ANF to either edges or node attributes in dataflow IR, and recover them when converting back.

In addition to the graph-level optimization, \system also incorporates operator-level optimizations.
Specifically, to support extensible backends so that both kernel libraries (e.g., cuBLAS and CUTLASS~\cite{cutlass}) and tensor compilation (e.g., TVM~\cite{chen2018tvm}) can be easily integrated, the dialect dispatching pass is used to determine which backend should be used for each operator.
The details are illustrated in \autoref{sec:impl_dialect}.
Besides, fusion passes (detailed in \autoref{sec:impl_fusion}) that fuse dialect operators together as closures are applied to reduce the kernel invocation and inter-operator communication overheads.



In the rest of this section, we present key system component designs in graph- and operator-level optimizations along with insights.

\subsection{Holistic Graph-Level Optimizations}



\subsubsection{Automatic Differentiation}
\label{sec:impl_autodiff}
In \system, we implement an automatic differentiation (Autodiff) pass that employs source code transformation with a closure-based method~\cite{pearlmutter2008reverse} to perform the differentiation .
Autodiff transforms each operator in forward graph by its corresponding adjoint function, which computes the derivative with respect to the inputs given the derivatives with respect to the outputs.
Specifically, each adjoint function returns the partial derivatives with respect to the inputs of the original function, as well as an ordered set of partial derivatives with respect to the outputs.

The advantage of self-contained Autodiff comes from flexible adjoint functions in terms of \textit{data dependency} and \textit{operator composition}. For data dependency, the adjoint function usually takes both primal operator's input $x$ and output $y$ as parameters to derive the gradient, which creates data dependency to both tensors in the forward graph.
For example, node \textbf{D} in \autoref{fig:train_workload} requires the output ($dy$) of node \textbf{C}, the input ($x$) and output ($y$) of node \textbf{A}.
However, for some operators, only $x$ or $y$ is sufficient to calculate the gradient.
Taking $y=tanh(x)$ as an example, gradient can be calculated by either $grad = 1-2\times tanh(x)$ or $grad = 1-2\times y$. As a result, we only require either $x$ or $y$ to calculate the gradient but not both, which not only reduces peak memory, but also increases fusion opportunity for the corresponding forward operators.

For operator composition, intuitively decomposing a backward operator to a series of small operators is able to provide more rooms for operator fusion. For example, $tanh\_dx(y) = 1-2\times y$ can be decomposed to a multiply followed by a subtract instead of an encapsulated \texttt{tanh\_dx} operator.
However, kernel libraries may outperform compilers for certain backward operators such as \texttt{Conv2D\_dx} even with the fusion opportunity sacrificed. With flexible adjoint functions, \system could select the best option for each case.

\subsubsection{Automatic Mixed Precision}
\label{sec:impl_autocast}
Although it is common to use half precision to support a larger batch size when training a gigantic mode, the time-to-train performance may still be inefficient due to the accuracy loss. Thus, modern DL frameworks offer a compromise solution that performs most computations in half precision; while keeping numerically sensitive computations in full precision.
This is called automatic mixed precision (AMP)~\cite{amp}.
However, whether an operator can be computed in half precision not only depends on its arithmetic expression but also the implementation.
As a result, taking a framework manipulated AMP graph may result in correctness issues for a DLC, as we will show in \autoref{sec:eval}.

\input{Figures/fig-cast}

In \system, we define our own list of numerical sensitive operators, and leverage our own AutoCast pass to insert \texttt{cast} operators to the graph for AMP training. In addition to guaranteeing the numerical correctness, AutoCast also optimizes the performance.
Specifically, although minimizing the number of inserted \texttt{cast} operators is straightforward, we observe that this is not always the best strategy.
Taking \autoref{fig:cast} as an example, if AutoCast minimizes the number of \texttt{cast} operators as in \autoref{fig:cast}b, then the shared \texttt{cast} cannot be fused into any of its consumers, as in \autoref{fig:cast}c.
In contrast, the exclusive \texttt{cast} operators in \autoref{fig:cast}d can be fused, as shown in \autoref{fig:cast}e.
Since the fused \texttt{cast} operation has almost no performance overhead, we could achieve better end to end performance accordingly.
The AutoCast pass in \system considers the operator fusion opportunities, and inserts exclusive \texttt{cast} operators to the consumer that is capable of fusing the element-wise predecessors and successors.

\subsubsection{Distribution Optimization}
\label{sec:impl_dist}

\system currently supports data parallelism\footnote{Model and pipeline parallelisms are ongoing future work.} to enable distributed training. The core idea is inspired by ZeRO~\cite{rajbhandari2020zero} that partitions optimizer state and gradients to reduce memory footprint on a single device when scaling out.
However, unlike DeepSpeed~\cite{rasley2020deepspeed} that manually implements ZeRO on top of PyTorch, \system automatically applies several transformation passes to analyze an IR graph for single device, and insert required collective operators to scale it out on multiple devices.

Compared to the manual implementation (i.e., DeepSpeed), our compiler-based solution introduces the following advantages.
First, the implemented transformation passes are generally applicable to all models and optimizers, significantly reducing the engineering efforts when a new model and optimizer emerge.
Second, it is easy to incorporate more advance optimizations. For example, when inserting collective operators for distributed training, we leverage CUDA streams to overlap computation and communication, so that inter-device communication overheads can be hidden.
In addition, we also apply horizontal fusion to collective operators to increase bandwidth utilization and reduce the communication overheads.
As we will demonstrate in \autoref{sec:eval}, by automatically applying these optimizations, \system is able to outperform DeepSpeed by $\sim14\%$ on a custom encoder model with 1.5 billion parameters on 8 NVIDIA A100 GPUs.

\subsubsection{Operator Scheduling}
\label{sec:impl_op_sche}

When converting a dataflow IR to ANF, any valid topological order is a valid execution order in terms of the correctness. However, certain execution orders may cause a much higher peak memory consumption which exceeds the DRAM capacity.
For instance, if we execute an operator that produces a large tensor too early, then we have to keep the large tensor occupy the memory for a long period of time.

To reduce the peak memory of an execution order, operator scheduling pass analyzes the memory footprint of each operator to determine when to execute it. Formally, let the total size of tensors produced by this operator be $p$, the total size of tensors that can be released after this operator be $c$, operator scheduler moves this operator forward when $p-c<0$, or backward when $p-c>0$. Thus, the operator that decreases memory consumption is executed as early as possible, and vice versa.

\subsubsection{Rematerialization}
\label{sec:impl_remat}
In the case that the peak memory exceeds the capacity after operator scheduling, another widely used optimization that is able to fit the model into the hardware is rematerialization~\cite{chen2016training}, which releases some intermediate tensors before encountering out of memory, and regenerates them later when needed.

Deep learning frameworks such as PyTorch rely on model developers to determine the checkpoints (i.e., the intermediate tensors that will not be regenerated) in forward, and replay every operators between two checkpoints in backward. This approach, however, brings two drawbacks. First, model developers have to manually identify checkpoints. Second, once checkpoints are enabled, even the memory budget is sufficient, every operator between two checkpoints must be replayed, which results in a significant performance overhead.

On the other hand, \system demonstrates that a compilation-based rematerialization is more usable and flexible. Specifically, the rematerialization pass in \system traverses an ANF IR and analyzes the memory footprint via liveness analysis.
To minimize rematerialization overheads, when the estimated memory consumption at a certain execution point exceeds the given budget, the rematerialization pass applies a cost function, which considers 1) the latency of recomputation, 2) the remaining use count, and 3) the size, to current alive intermediate tensors.
It then identifies one or more intermediate tensors, and breaks their liveness to two intervals, where the second interval will be initialized by replaying the corresponding operator.
Consequently, this approach brings two benefits. First, since this process is fully automatic, developers do not need to insert checkpoints to the model. Second, \system only rematerializes necessary tensors to fit into the memory budget, so the performance overhead is moderated compared to manual checkpoints.

\subsection{Operator-Level Optimizations}

\subsubsection{Operator Dialects}
\label{sec:impl_dialect}

One platform may have multiple kernel implementations, but there is no single library/compiler that outperforms others in all operators.
For example, according to our evaluations on NVIDIA V100 GPU, although both cuDNN and TVM provide certain CUDA kernels for NVIDIA GPUs, cuDNN achieves 1.8$\times$ speedups over TVM in Conv2D backward, but TVM is able to achieve up to 100$\times$ speedups (with auto-tuning) in softmax backward kernel.
As a result, it is inevitable to have a flexible kernel selection mechanism to achieve the optimal performance.
Even worse, more and more backend implementations now introduce the capability of operator fusion such as cuDNN v8 \cite{cudnnv8} and CUTLASS \cite{cutlass} while supporting different fusion patterns, so it is even more challenging as described in \cite{jeon2021collage}.

Motivated by MLIR~\cite{lattner2021mlir} that introduces IR dialects, we introduce the operator dialects to address this issue. Similar to the dialect language in MLIR, we register each backend implementation as a new dialect in the IR, e.g., ``cudnn'',  ``tvm'', etc., and allow each dialect to have customized operator attributes.
Different from MLIR which views all backends as dialects, \system comes with a set of pre-defined operators as base operators. This can reduce the efforts of adding a new dialect operator by sharing certain operator attributes from base operators such as type relation function. 
In addition, users do not need to write the code for translation between every dialect pair as MLIR requires.
Nonetheless, it still provides the flexibility that allows dialect operators to have customized semantics via attribute tables, as shown in \autoref{fig:dialect-op-attr}, and more importantly, enables different fusion patterns for each dialect (detailed in \autoref{sec:impl_fusion}).

\input{Figures/fig-dialect-op}

In \autoref{fig:dialect-op}, a dialect operator is registered to a base operator with a dispatching priority, which can be pre-defined by developers based on prior experiences or derived from profiling results. A dialect operator is also associated with a dialect. For example, \texttt{tvm.conv2d} that belongs to dialect \texttt{tvm} is registered to the base operator \texttt{conv2d} with priority 10. In addition, a dialect is allowed to be only enabled on certain devices. For instance, dialect \texttt{tvm} can be used on both CPU and GPU; while dialect \texttt{cudnn} and \texttt{cutlass} are only available for GPU. If a dialect is disabled in \system, all dialect operators in this dialect will not be included in the IR or dispatched at runtime.

\input{Figures/fig-dialect-op-attr}

Every operator including both base and dialect operators has its own attribute table. The attribute table of a base operator contains common attributes that share across its dialect operators, such as argument schema, type relation functions, etc., as shown in \autoref{fig:dialect-op-attr}. Most attributes of dialect operators can inherit from base operators, so that it significantly reduces the developer effort to add new dialect operators (blue items in \autoref{fig:dialect-op-attr}). On the other hand, dialect operators can overwrite certain attribute value on top of the base operators and have its own specific attributes (red and green items respectively). For example, in \autoref{fig:dialect-op-attr}, \texttt{cudnn.dropout} has a different type relation function as the cuDNN implementation needs to reserve extra space for the states.

\subsubsection{Operator Fusion}
\label{sec:impl_fusion}

Operator fusion combines multiple operators into a single kernel and can greatly reduce kernel execution time by avoiding the data transfer of intermediate results between cache and main memory.
It is particularly effective for GPUs and specialized accelerators.
As we mentioned before, multiple backends have the capabilities for operator fusion and their fusion patterns are different from each other.
For example, a template based library CUTLASS~\cite{cutlass} is capable of fusing a general matrix multiplication (GEMM) and its subsequent element-wise activation functions.
On the other hand, as a deep learning compiler, Apache TVM fuses operators based on their fusion rules.
Although TVM may not achieve better performance than the handcrafted kernels, in some cases, it is able to achieve better end-to-end performance by effectively fusing operators in a way that vendor libraries do not cover, as reported in \cite{chen2018tvm}.
In \system, we aim to have a fusion plan that benefits from all backends.
For example, in this particular case, we prefer to choose the CUTLASS one when the pattern is matched due to its handcrafted high performance; while the rest operators can be fused and optimized by TVM.

To accommodate different levels of fusion capabilities for all backends, \system divides the operator fusion optimization into two categories: {\em pattern-based fusion} and {\em rule-based fusion}.
For example, fusion for CUTLASS and cuDNN backends belong to pattern-based category as they only support limited fusion patterns (e.g., general matrix multiplication (GEMM) followed by an element-wise activation). 
On the other hand, TVM belongs to rule-based fusion as it performs general-purpose operator fusion based on fusion rules \cite{chen2018tvm}.

\input{Tables/tbl-fusion-pattern}
\input{Figures/fig-fusion}

For pattern-based fusion, \system provides an interface for developers to register fusion patterns. 
Thanks to operator dialect described in \autoref{sec:impl_dialect}, we can attribute each pattern to a specific dialect, to indicate which backend is used for code generation of the fused operators. 
Each pattern is associated with a priority, which determines the order of application of these fusion patterns.
We also set priorities to rule-based fusion, and the operator fusion pass applies these fusion patterns and fusion rules one at a time based on their priorities in a descending order.
The orders of pattern matching will lead to different fused graphs and can have different performance.
Although the priorities are now set by developers based on heuristics, we plan to derive the priorities from profiling results and perform search on all possible graph partition solutions based on the fusion patterns and fusion rules to find optimal fusion plans.

We use an example in \autoref{fig:fusion} to illustrate how operator fusion takes action and applies the fusion patterns and rules.
\autoref{tbl:fusion-pattern} gives an example of a few fusion patterns related to \texttt{conv2d} and TVM fusion rules with priorities.
First, the CUTLASS fusion pattern for float16 data type in the fusion table has the highest priority.
The optimization pass finds a matching in the program and then replaces the matched sub-graph by a fused operator with dialect \texttt{cutlass} and a function call.
The IR is then transformed from \autoref{fig:unfused} to \autoref{fig:fuse-cutlass}.
After that, no patterns other than TVM fusion rules can be found applicable to the transformed program.
\autoref{fig:fuse-tvm} shows the final program after the fusion optimization.

%% file: Figures/fig-overview.tex
\begin{figure}[!tb]
	\centering
	\includegraphics[width=0.8\linewidth]{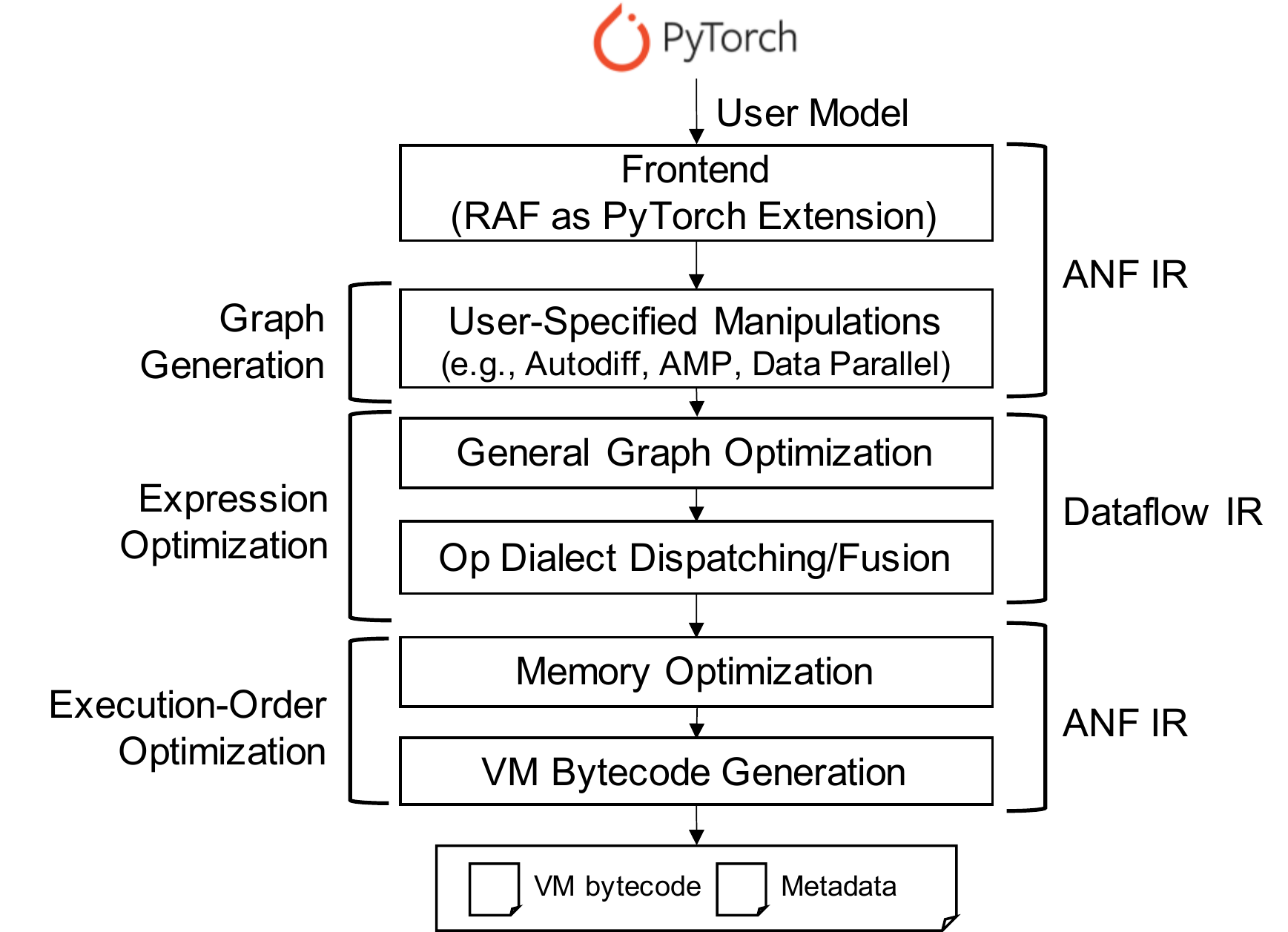}
	\caption{\system compilation flow.}
	\label{fig:overview}
\end{figure}

%% file: Figures/fig-cast.tex
\begin{figure}[!tb]
	\centering
	\includegraphics[width=0.85\linewidth]{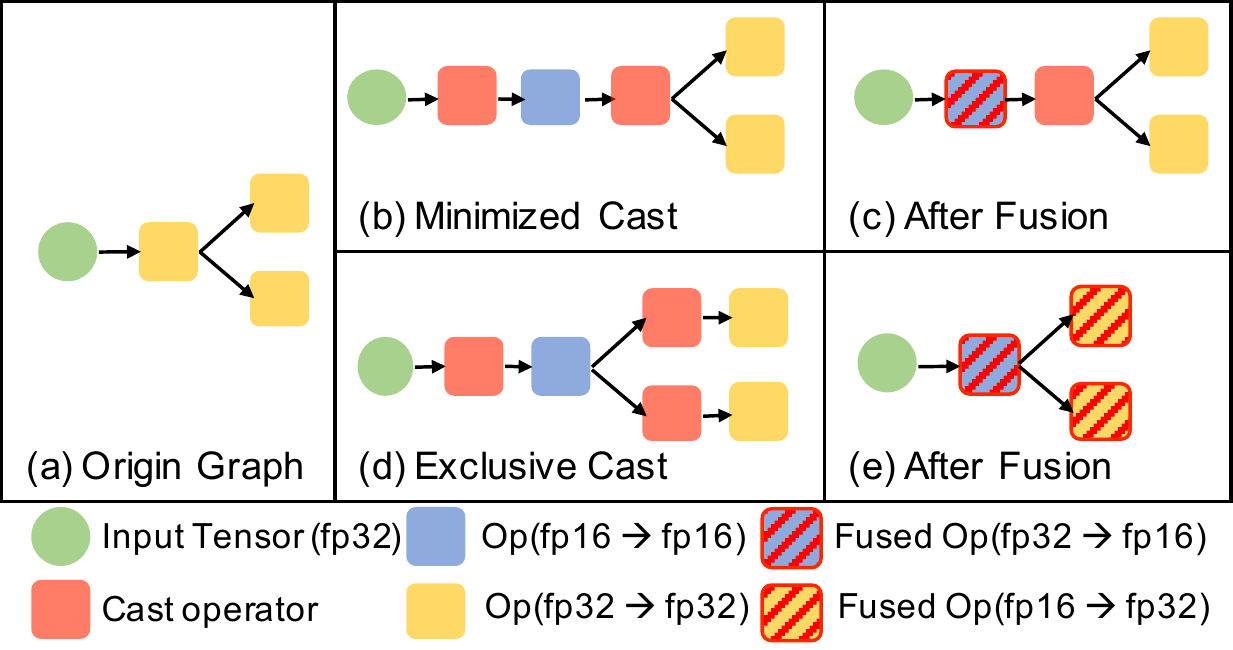}
	\caption{Impact of inserting minimized or exclusive cast operators for AMP on fusion.}
	\label{fig:cast}
\end{figure}

%% file: Figures/fig-dialect-op.tex
\begin{figure}[t]
    \centering
    \includegraphics[width=\linewidth]{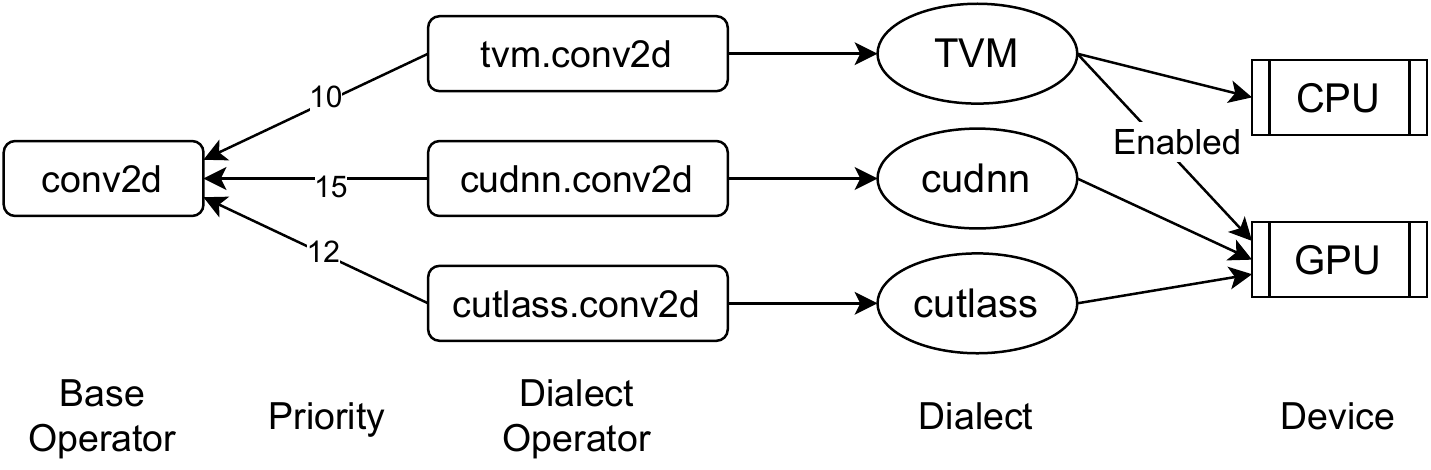}
    \caption{A dialect operator associated with one dialect is registered to a base operator with a dispatching priority. A dialect is only enabled on certain devices.}
    \label{fig:dialect-op}
\end{figure}

%% file: Figures/fig-dialect-op-attr.tex
\begin{figure}[t]
    \centering
    \includegraphics[width=\linewidth]{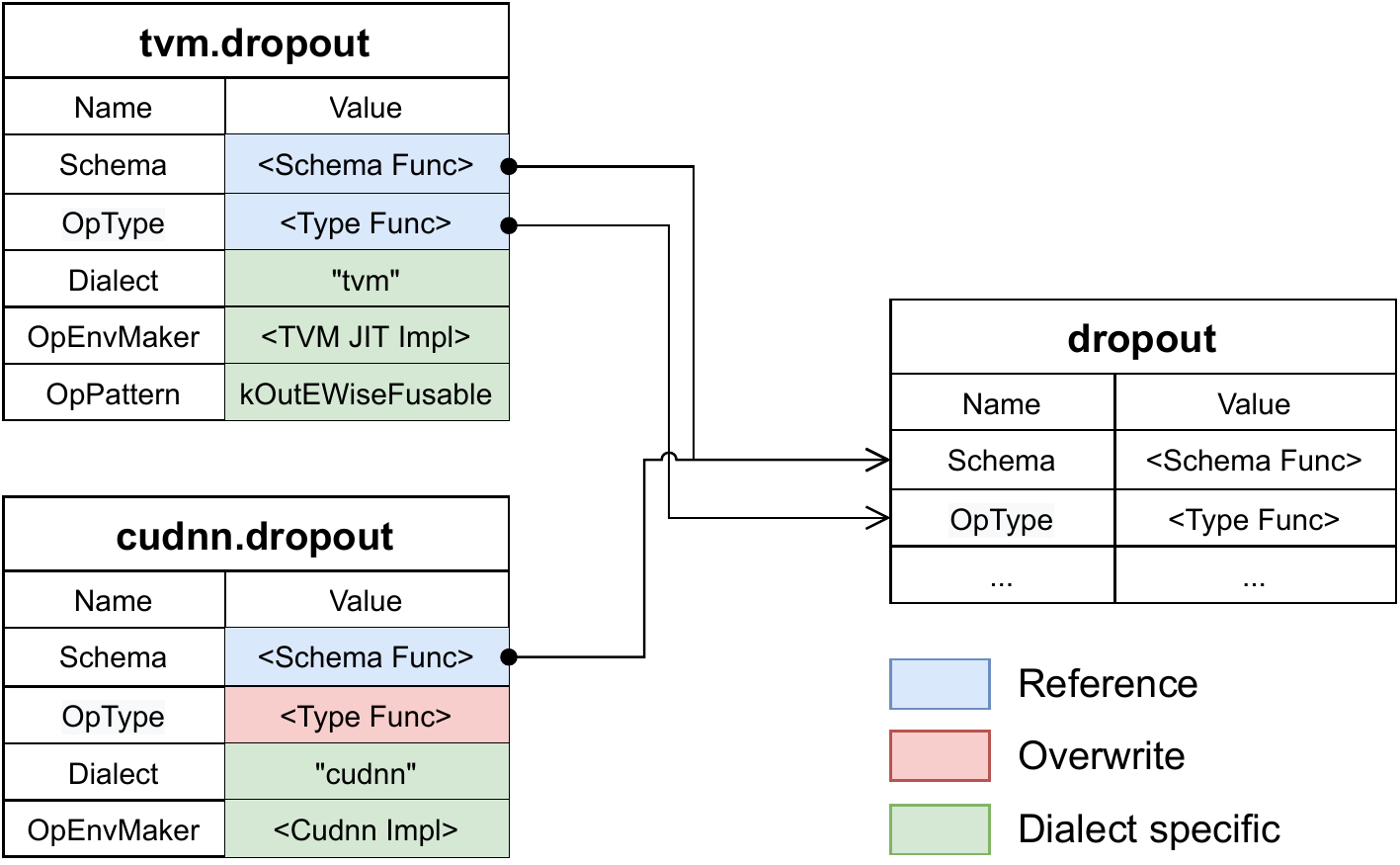}
    \caption{The attribute table for base and dialect operators. Attributes from base operators can be inherited (blue), overridden (red), and customized (green).}
    \label{fig:dialect-op-attr}
\end{figure}

%% file: Tables/tbl-fusion-pattern.tex
\begin{table}[t]
\centering
\begin{tabular}{c|c|c}
    \toprule
    Pattern / Rules & Dialect & Priority \\ \midrule
    conv2d + bias + relu (fp16) & cutlass & 15 \\
    conv2d + bias + relu & cudnn (>=8) & 12 \\
    conv2d & cudnn & 10 \\
    conv2d + bias + relu & cutlass & 8 \\
    TVM fusion rules & tvm & 5 \\
    \bottomrule
\end{tabular}
\caption{Fusion patterns and rules with priorities.}
\label{tbl:fusion-pattern}
\end{table}

%% file: Figures/fig-fusion.tex
\newsavebox{\unfusedir}
\begin{lrbox}{\unfusedir}
\begin{lstlisting}
%1 = raf.op.conv2d(%x: f16, %w: f16);
%2 = raf.op.relu(%1);
%3 = raf.op.add(%2, %y);
%4 = raf.op.multiply(%3, %z);
\end{lstlisting}
\end{lrbox}

\newsavebox{\fusecutlassir}
\begin{lrbox}{\fusecutlassir}
\begin{lstlisting}
%1 = fn(%p0, %p1, dialect="cutlass") {
  %2 = raf.op.cutlass.conv2d(%p0, %p1);
  %3 = raf.op.cutlass.relu(%2)
};
%4 = %1(%x: f16, %w: f16);
%5 = raf.op.add(%4, %y);
%6 = raf.op.multiply(%5, %z);
\end{lstlisting}
\end{lrbox}

\newsavebox{\fusetvmir}
\begin{lrbox}{\fusetvmir}
\begin{lstlisting}
%1 = fn(%p0, %p1, dialect="cutlass") {
  %2 = raf.op.cutlass.conv2d(%p0, %p1);
  %3 = raf.op.cutlass.relu(%2)
};
%4 = %1(%x: f16, %w: f16);
%5 = fn(%p0, %p1, %p2, dialect="tvm", primitive=1) {
  %6 = raf.op.tvm.add(%p0, %p1);
  %7 = raf.op.tvm.multiply(%6, %p2)
};
%8 = %5(%4, %y, %z);
\end{lstlisting}
\end{lrbox}

\begin{figure}[!h]
    \centering
    \subfloat[][Before fusion.]{
        \usebox{\unfusedir}
        \label{fig:unfused}
    }
    \qquad
    \subfloat[][After applying CUTLASS fusion pattern.]{
        \usebox{\fusecutlassir}
        \label{fig:fuse-cutlass}
    }
    \qquad
    \subfloat[][After applying TVM fusion rules.]{
        \usebox{\fusetvmir}
        \label{fig:fuse-tvm}
    }
    \caption{An example IR for fusion illustration.}
    \label{fig:fusion}
\end{figure}

%% file: Sections/4.impl.tex
\section{Implementations}

In addition to the compiler design presented in the previous section, this section introduces the frontend as well as the runtime of \system, making it an end-to-end deep learning system.

\subsection{\system as PyTorch Extension}
\label{sec:impl_frontend}

We observe that users tend to use the interactive interface provided by PyTorch~\cite{paszke2019pytorch} to write and tune the deep learning models for training.
As a result, we aim to integrate \system with PyTorch user interface and programming paradigm.
However, it is challenging because PyTorch embraces an imperative style programming paradigm to provide the best usability, which dynamically interprets and executes a model graph operator by operator.
On the other hand, compilation-based frameworks such as \system intend to compile an entire model graph at once to apply graph-level optimizations before execution for better performance.
This programming paradigm deviates drastically from the one PyTorch adopts and makes it difficult for users to print intermediate results or make modification to their models.
Consequently, it will take huge amount of work to run their existing models or training scripts with \system.

To minimize the user efforts of enabling \system in their training jobs, \system incorporates Lazy Tensor Core (LTC)~\cite{suhan2021lazytensor} as its imperative frontend.
The frontend is designed as an extension to PyTorch and can be easily plugged in officially released PyTorch packages without changing upstream PyTorch codebase. 
The core idea of LTC is to register a new backend in PyTorch. Unlike operators in other backends that generate output tensor when executed, operators in LTC generate a \textit{lazy tensor}, which is only a symbol that connects to its input lazy tensors.
In this way, when users execute a PyTorch model imperatively, no execution is actually performed but only the operators are traced and recorded, so that the complete model graph is obtained.
The actual compilation and execution happen when the LTC synchronization API is invoked.

\input{Figures/fig-code}

Figure~\ref{fig:code} shows an example of enabling \system via LTC for a PyTorch model by changing 2-3 lines of code. Specifically, at L4, a user moves the model from CPU to \textit{lazy} device, in which backend is registered to be \system when importing \system.
Then, a \system specific API at L7 has to be invoked to wrap the forward model. This helps us to capture a complete model structure including control flow.
With above changes, the model is now in the lazy mode. It means that during L9-12, the operators are \textit{not} executed at the time of visiting, but just staged as a symbol in \system IR form a graph.
The collected IR graph will then be optimized, compiled and executed at the synchronization point users created at L13.

\subsection{Runtime Virtual Machine}
After the model graph IR has been optimized, we compile the IR to the virtual machine (VM) bytecode that could be executed by \system VM runtime.
We choose virtual machine instead of graph interpreter because of the following reasons. First, it can be extended to support non-DAG executions (e.g., the model with control flow). Second, it is able to dynamically manage memory for dynamic shapes.


The VM execution flow is described as follows.
Before kicking off the training loop, \system initializes a VM instance to load bytecode, initialize registers and memory pool, etc.
For each training iteration, the VM interprets the bytecode to manipulate registers, manage memory pool, and invoke the corresponding kernels.
In particular, when an operator or a closure (i.e., a fused operator) is visited for the first time, VM invokes a particular backend, which is determined by dialect dispatching pass during IR transformation, to perform just-in-time (JIT) compilation to generate and cache the executable binary.
We will demonstrate in \autoref{sec:eval} that since JIT compilation are one-time overheads,
it will not hurt the user experience as training a model usually needs thousands to millions iterations.

%% file: Figures/fig-code.tex
\begin{figure}[t]
  \centering
    \begin{lstlisting}[language=python,breaklines=false,escapechar=!]
import torch, raf
from torch.optim import Adam
model = # initialize a torch model
!\colorbox{backcolor}{model = model.to("lazy")}!
model.train()
optimizer = Adam(model.parameters())
!\colorbox{backcolor}{model = raf.jit.script(model)}!
for inp, label in data_loader():
  out = model(inp.to("lazy"))
  loss = compute_loss(out, label.to("lazy"))
  out.backward()
  optimizer.step()
!\colorbox{backcolor}{\ \ raf.mark\_step()}!
    \end{lstlisting}
    \caption{Code snippet of \system frontend programming model.}
    \label{fig:code}
\end{figure}

%% file: Sections/5.evaluation.tex
\section{Evaluation}
\label{sec:eval}

In this section, we evaluate \system with popular transformer models -- BERT~\cite{devlin2018bert}, RoBERTa~\cite{liu2019roberta} and GPT-2~\cite{radford2019language} -- on Amazon EC2 GPU instances. The models are pulled from Huggingface transformer library~\cite{wolf2019huggingface} without modification.

The baselines we choose for single device training are eager mode and torchscript in PyTorch 1.12.1 with CUDA 11.3.
We also choose DeepSpeed 0.5.10 for distributed training evaluation.
In addition, we also evaluate \system against PyTorch/XLA 1.12+8baddab (released at 05/09/2022), because it also leverages Lazy Tensor Core (LTC)~\cite{suhan2021lazytensor} to lower PyTorch model to XLA for compilation.
Note that since JAX~\cite{jax2018github} does not support automatic mixed precision (AMP), and its underlying compile engine is also XLA, we bypass the JAX in our evaluation and leverage the results of PyTorch/XLA for evaluation and analysis.
As a result, it would be straightforward to illustrate that 1) compilation techniques could help improve training efficiency, and 2) self-contained backward graphs as well as extensible backends could be the key to achieve better performance.

\subsection{Throughput Evaluation}

\input{Figures/fig-exp-thrpt}

We first evaluate the throughputs of processing a mini-batch to answer the following questions:

\begin{enumerate}
    \item What speedup can \system achieve over PyTorch and PyTorch/XLA?
    \item How holistic compilation optimization could help achieve higher throughputs?
\end{enumerate}

We perform this evaluation on a single NVIDIA V100 GPU with 16 GB on-device DRAM. We use WikiText-2~\cite{merity2016pointer} as the training dataset. The models and corresponding tokenizers are from Huggingface~\cite{wolf2019huggingface} with sequence length set to 512.
We use Adam optimizer~\cite{kingma2014adam} implemented in PyTorch with learning rate 1e-5 and eps 1e-6.

\autoref{fig:thrpt} depicts the throughputs\footnote{The one-time compilation overhead is not included, which will be discussed in \autoref{sec:eval:time_to_train}.} over batch sizes for all models. For each model, we evaluate the throughput of training with full precision (left) and float16 automatic mixed precision (AMP)\footnote{AMP is able to achieve better convergence than half precision, so it is more preferable to train a model with AMP when the model can be fit into a single device.} (right). As can be seen, \system achieves higher batch size and throughput than PyTorch eager mode. We do not enable user-defined gradient checkpointing in PyTorch models, because it constantly introduces $\sim2\times$ latency overhead even when the memory is sufficient.

Meanwhile, we observe that \system could support larger batch sizes than PyTorch in both eager and torchscript mode when user-defined gradient checkpointing is disabled. This is because \system analyzes the memory footprint of the model, and automatically invokes rematerialization (\autoref{sec:impl_remat}) to release some intermediate tensors in forward pass and re-compute them in backward propagation.
Since \system only re-computes a minimal number of tensors, the latency overhead is moderated in most cases, so we could still observe throughput improvement, such as BERT-large, RoBERTa-base and RoBERTa-large.
On the other hand, we also observe that for a few models such as BERT-base AMP and GPT-2, the throughput drops with the maximum batch size. This is because the peak memory of these models is substantially larger than 16~GB, the GPU DRAM size, under large batch size. To satisfy the memory constraint, \system has to re-compute many more tensors, which moderates the throughput improvement as the batch size grows.

Besides, we can see that \system outperforms PyTorch/XLA in large models (BERT-large with 340 million parameters and RoBERTa-large with 355 million parameters). This again illustrates that getting control of the entire procedure of generating and executing training graph of the model is capable of achieving better performance.
On the other hand, we ignored the results of PyTorch/XLA in GPT-2 in \autoref{fig:thrpt} because it failed to generate the correct results, which will be verified in the next subsection.

\input{Figures/Fig-exp-ablation-raf}

To better understand how operator fusion and memory optimization contribute to the training throughput, we conduct an ablation study with BERT-large model in \autoref{fig:ablation_raf}. All experiments were done with the batch size leading to the best throughput. As can be seen, \system-origin even performs worse than PyTorch without any optimizations due to various overheads (e.g., kernel launching, memory allocation, etc.).
By getting rid of these overheads with operator fusion, \system achieves 30\% speedup.
Furthermore, increasing the batch size from 4 to 6 will trigger rematerialization with minimal overhead (2.5\% more latency to recompute 181 operators), which brings another 10\% more throughput due to better GPU utilization.


\subsection{Time-to-Train Performance Evaluation}
\label{sec:eval:time_to_train}

In this evaluation, we attempt to answer two questions:

\begin{enumerate}
    \item Whether \system delivers the numerical correct results and have the same convergence capability as the state-of-the-art (i.e., PyTorch)?
    \item How the one-time compilation overhead is moderated when training large models?
\end{enumerate}

We design the following experiments to answer these questions: Although MLPerf~\cite{mattson2020mlperf} suggests to perform a complete pre-training from scratch and measure the time to achieve the target metric as the time-to-train performance, it takes a long time on multiple-device-multiple-node platforms and is not suitable for our evaluation.
For the sake of time, we perform \textit{continue pre-training} based on pretrained models from Huggingface transformer library. Specifically, we first train the model using native PyTorch for 10 epochs, and use the final loss from PyTorch at 10th epoch as the \textit{target loss} to train \system and PyTorch/XLA. In other words, we train \system and PyTorch/XLA for required epochs until they reach the loss PyTorch achieved at the 10th epoch.
Note that for all training tasks, we use the same random seed and hyper-parameters for all training tasks for apples-to-apples comparison. This is reasonable because both \system and PyTorch/XLA aim to seamlessly support PyTorch models and programming paradigm.

\input{Figures/fig-exp-train}

The loss trend of all three frameworks are depicted in \autoref{fig:train}. Note that since \system performs holistic optimizations, the IRs with backward, optimizer, and AMP between PyTorch and \system are not exactly the same, which result in different convergence trend. However, as can be seen, \system spends the same number of epochs as PyTorch to achieve the target loss (1.128 for BERT-large and 1.458 for GPT-2), proving that \system preserves the numerically correctness.

On the other hand, we can see from \autoref{fig:train} that although PyTorch/XLA achieves the target loss as PyTorch on BERT-large, it fails on GPT-2 and the loss fluctuates over time. Deep learning practitioners usually adjust the hyper-parameters such as learning rate and momentum to avoid this issue, but since we perform continue pre-training based on trained parameters, and both PyTorch and \system could deliver smooth loss trends, we believe PyTorch/XLA should be capable of training the models with the same hyper-parameters. As a result, it is likely due to the fact that PyTorch/XLA fails to preserve the numerically correctness when optimizing GPT-2.

\input{Tables/tbl-exp-train}

Meanwhile, in \autoref{tbl:exp-train}, we report the time-to-train along with model setup time (i.e., one time compilation) in minutes.
Although \system has a one-time compilation overhead compared to PyTorch, it still achieves the target loss in a shorter time. 
Specifically, given the elapsed time of the first epoch $t_0$, which includes compilation overheads, and the average elapsed time of the rest epochs $t_a$, we could estimate the total training time after $N$ epochs: $T=t_0+(N-1) \times t_a$. Given that the average elapsed time for training an epoch for BERT-large and GPT2 on PyTorch and \system are 178 vs. 130 and 75 vs. 57 seconds, respectively, \system achieves better time-to-train performance than PyTorch when $N$ is larger than 8 and 9 for BERT-large and GPT-2, respectively. Consequently, \system is capable of delivering better time-to-train performance than PyTorch for large language models, as their pre-training needs many more epochs to converge.

\subsection{Data Parallelism Evaluation}

\input{Figures/fig-exp-dist}

We use \autoref{fig:dist} to demonstrate large model training with data parallelism. The baseline is DeepSpeed~\cite{rasley2020deepspeed}\footnote{Data parallelism support in PyTorch/XLA is not stable yet, so we skip it in this evaluation.}, the ZeRO~\cite{rajbhandari2020zero} implementation on top of PyTorch. In this experiment, we evaluate the throughput of training a proprietary custom transformer model with 1.5B parameters on Amazon~EC2~p4d instance with 8 NVIDIA A100 GPUs and 40~GB DRAM each. As can be seen, \system achieves $\sim14\%$ throughput improvements due to the optimizations presented in \autoref{sec:impl_dist}.

\input{Figures/fig-exp-ablation-dist}

To further illustrate the optimizations mentioned in \autoref{sec:impl_dist}, we conduct an ablation study of the above experiment, as shown in \autoref{fig:ablation_dist}. 
In addition, \autoref{fig:dist_profile} depicts a detail breakdown of per-iteration execution.
As can be seen, by overlapping the computation and communication using CUDA streams \system is able to match the DeepSpeed manual optimized throughput.
Moreover, we can see from \autoref{fig:dist_profile} that the communication in \textit{Overlapping} is composed by a huge number of collective operators, which results in a long latency and synchronization overheads.
Accordingly, after enabling horizontal fusion that aggregates collective operators such as all-gathers and reduce-scatters, the profiling of \textit{Fusion} shows that the communication latency is significantly reduced and can be completely hidden.
It shows that an automatic compiler solution is capable of achieving the same or better performance against manual optimization at the framework level.

\input{Figures/fig-exp-profile-dist}

%% file: Figures/fig-exp-thrpt.tex
\begin{figure}[!ht]
    \centering
    \subfloat[][BERT-base-uncased]{
        \includegraphics[width=0.43\textwidth]{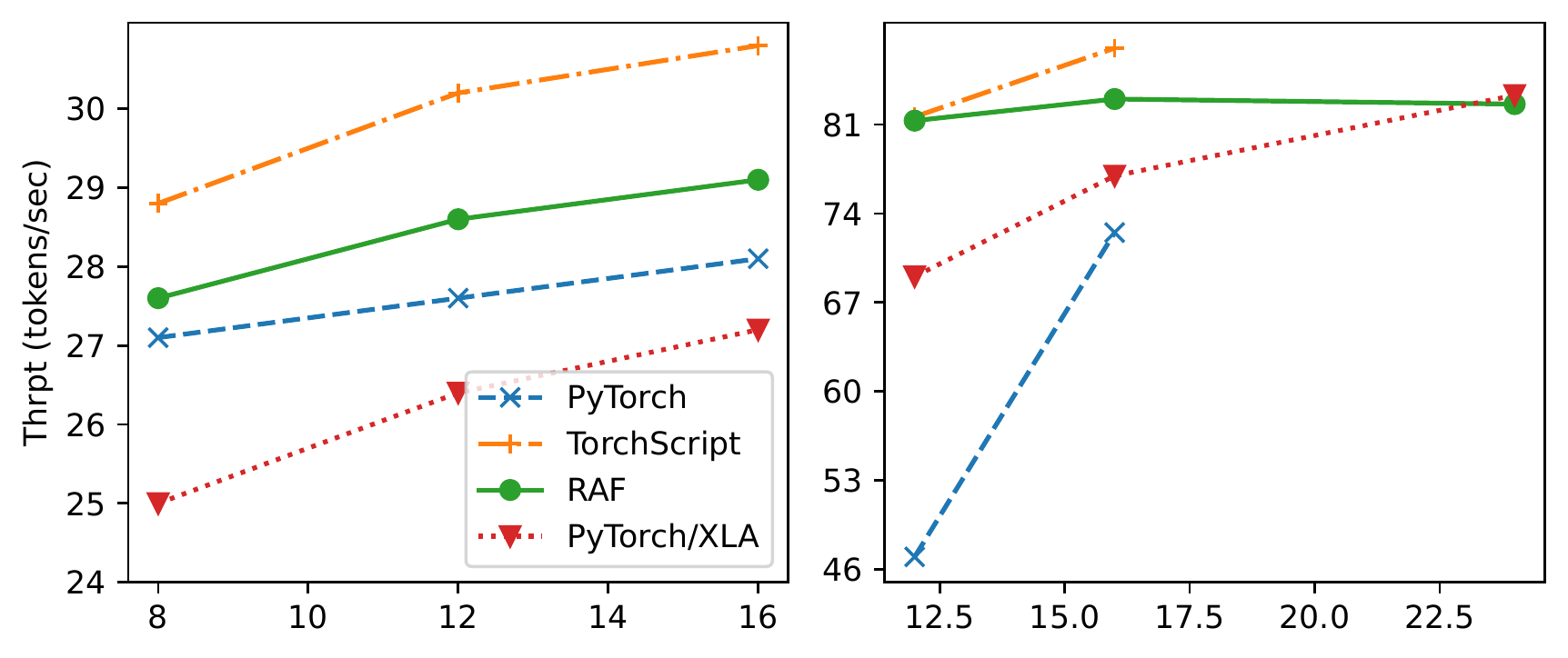}
        \label{fig:thrpt-bert-base}
    }
    \qquad
    \subfloat[][BERT-large-uncased]{
        \centering
         \includegraphics[width=0.43\textwidth]{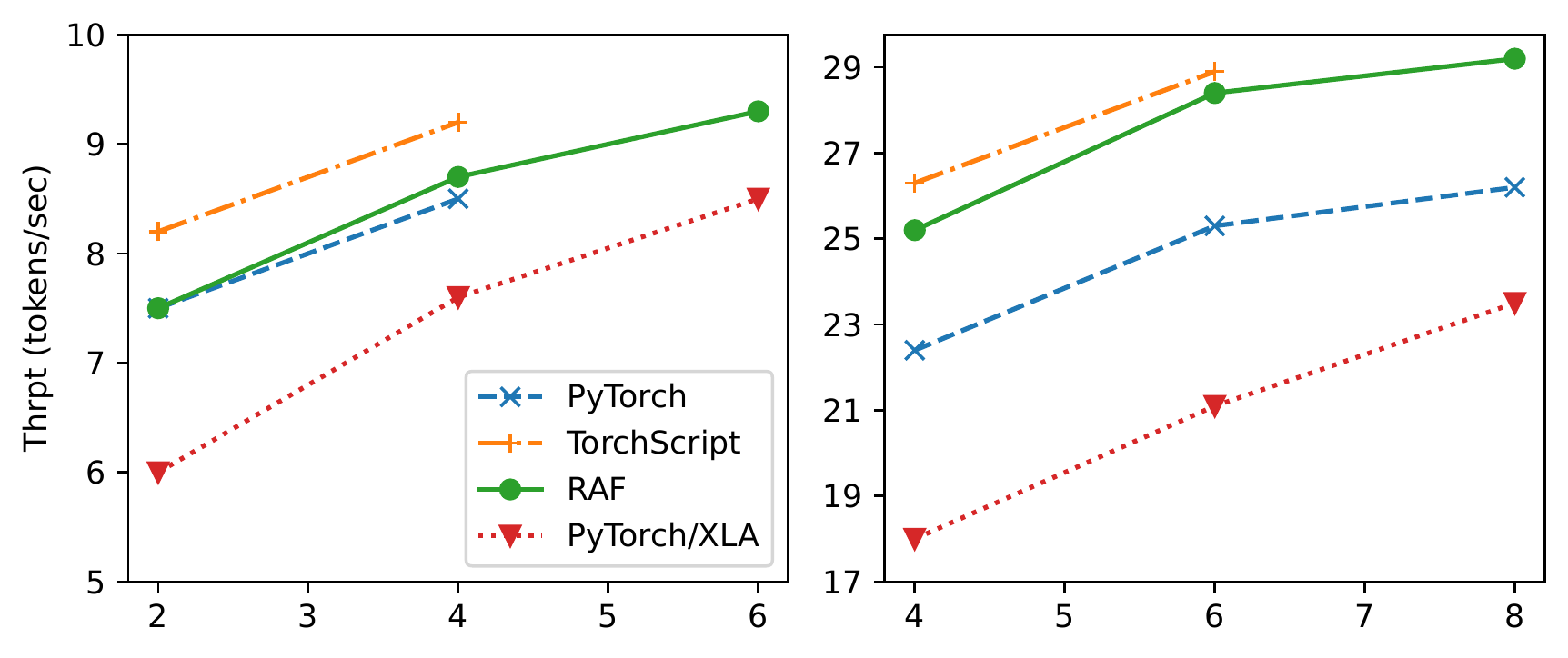}
        \label{fig:thrpt-bert-large}
    }
    \qquad
    \subfloat[][RoBERTa-base]{
        \centering
         \includegraphics[width=0.43\textwidth]{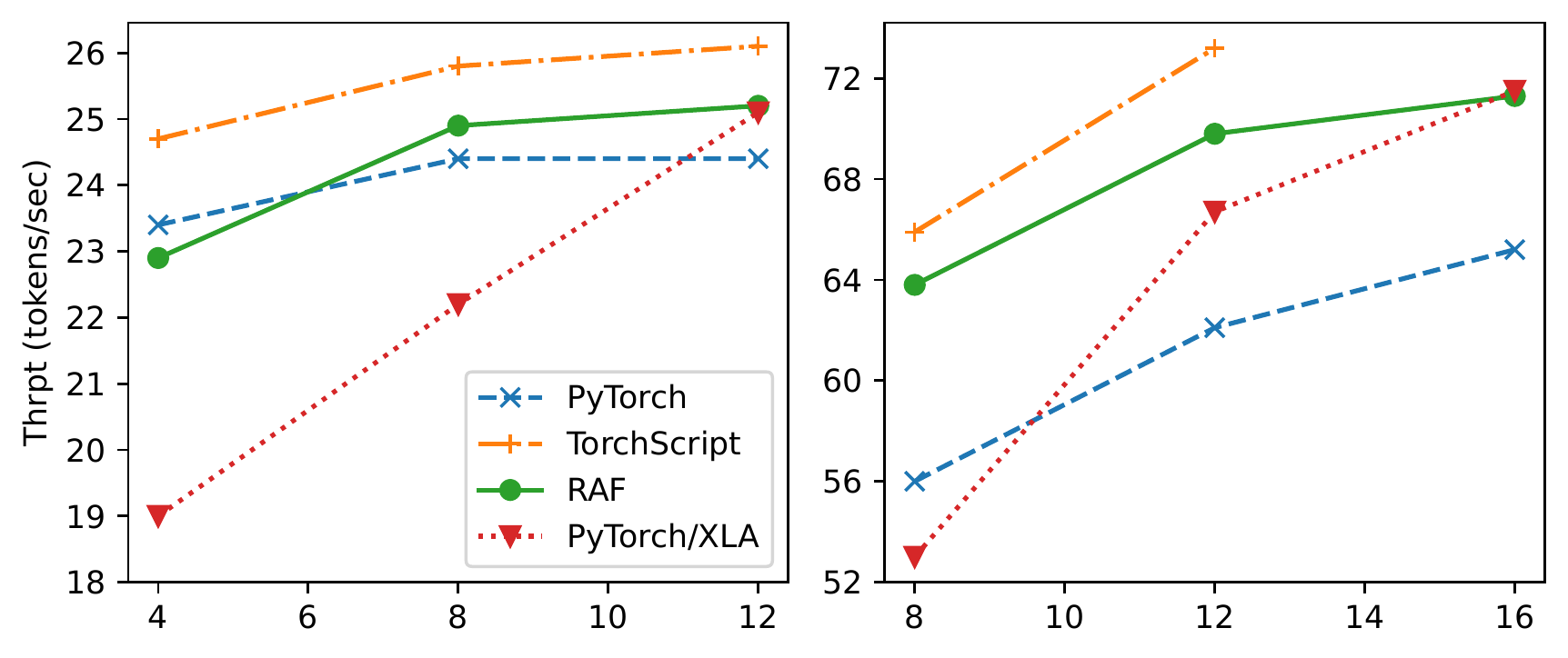}
        \label{fig:thrpt-roberta-base}
    }
    \qquad
    \subfloat[][RoBERTa-large]{
        \centering
         \includegraphics[width=0.43\textwidth]{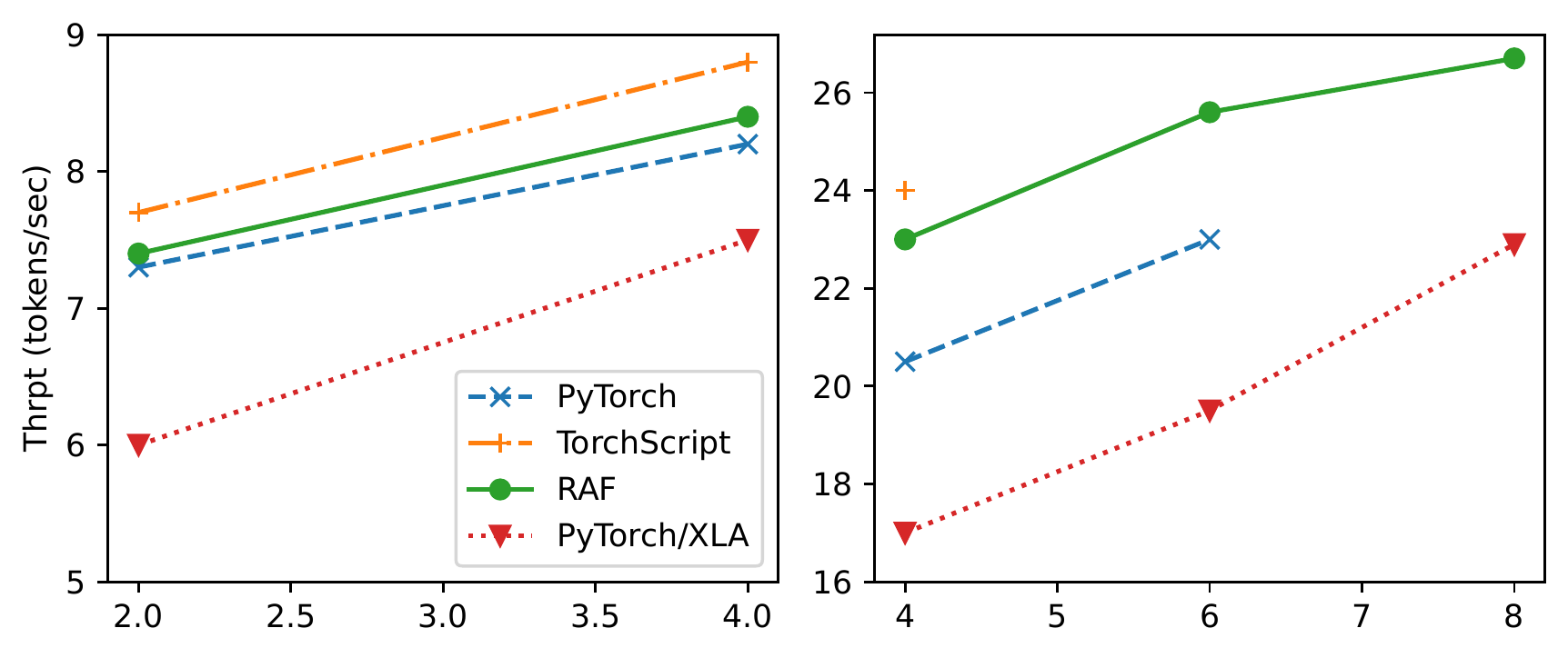}
        \label{fig:thrpt-roberta-large}
    }
    \qquad
    \subfloat[][GPT-2]{
        \centering
        \includegraphics[width=0.43\textwidth]{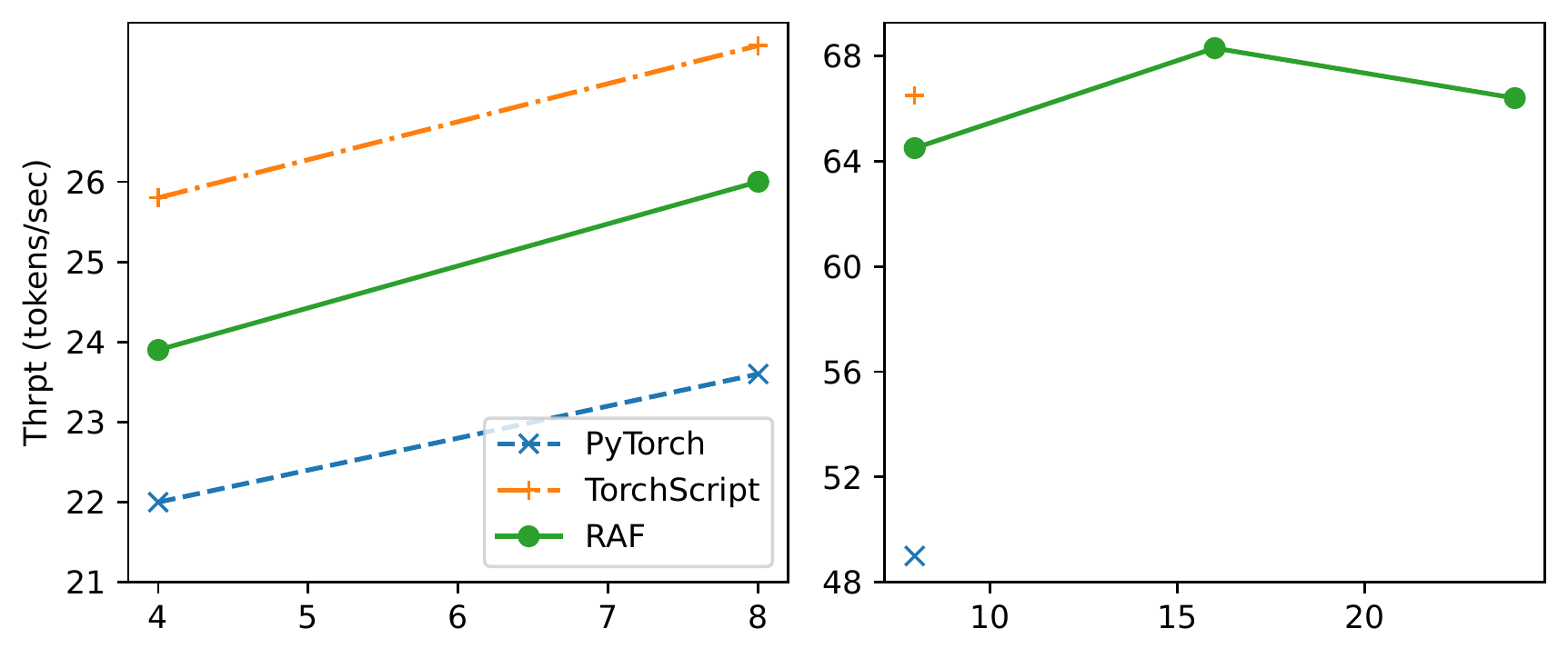}
        \label{fig:thrpt-gpt2}
    }
    \caption{Throughput (y-axis) vs. batch size (x-axis). Missing points indicate out-of-memory. Results of GPT-2 with PyTorch/XLA is ignored due to incorrect results.}
    \label{fig:thrpt}
\end{figure}

%% file: Figures/fig-exp-ablation-raf.tex
\begin{figure}[tbh]
    \centering
    \includegraphics[width=0.47\textwidth]{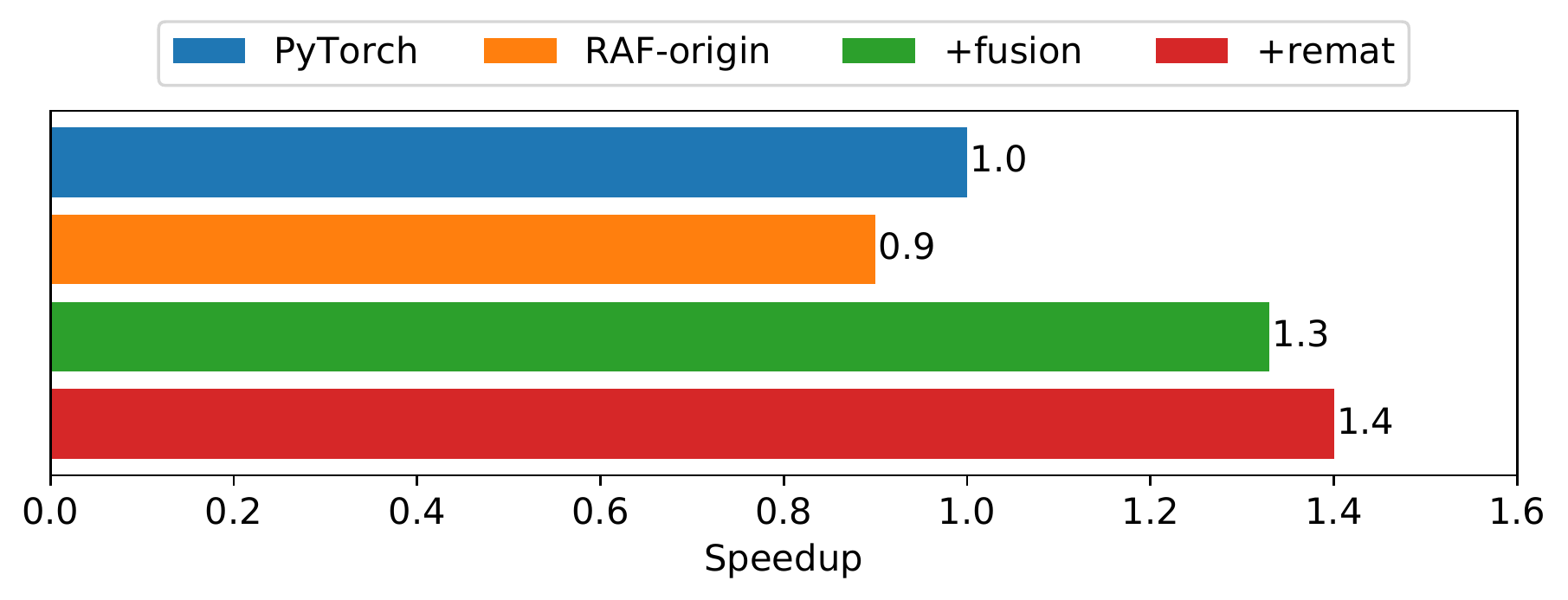}
    \caption{Ablation study of RAF with BERT-large-uncased compared to PyTorch.}
    \label{fig:ablation_raf}
\end{figure}

%% file: Figures/fig-exp-train.tex
\begin{figure}[tbh]
    \centering
    \includegraphics[width=0.47\textwidth]{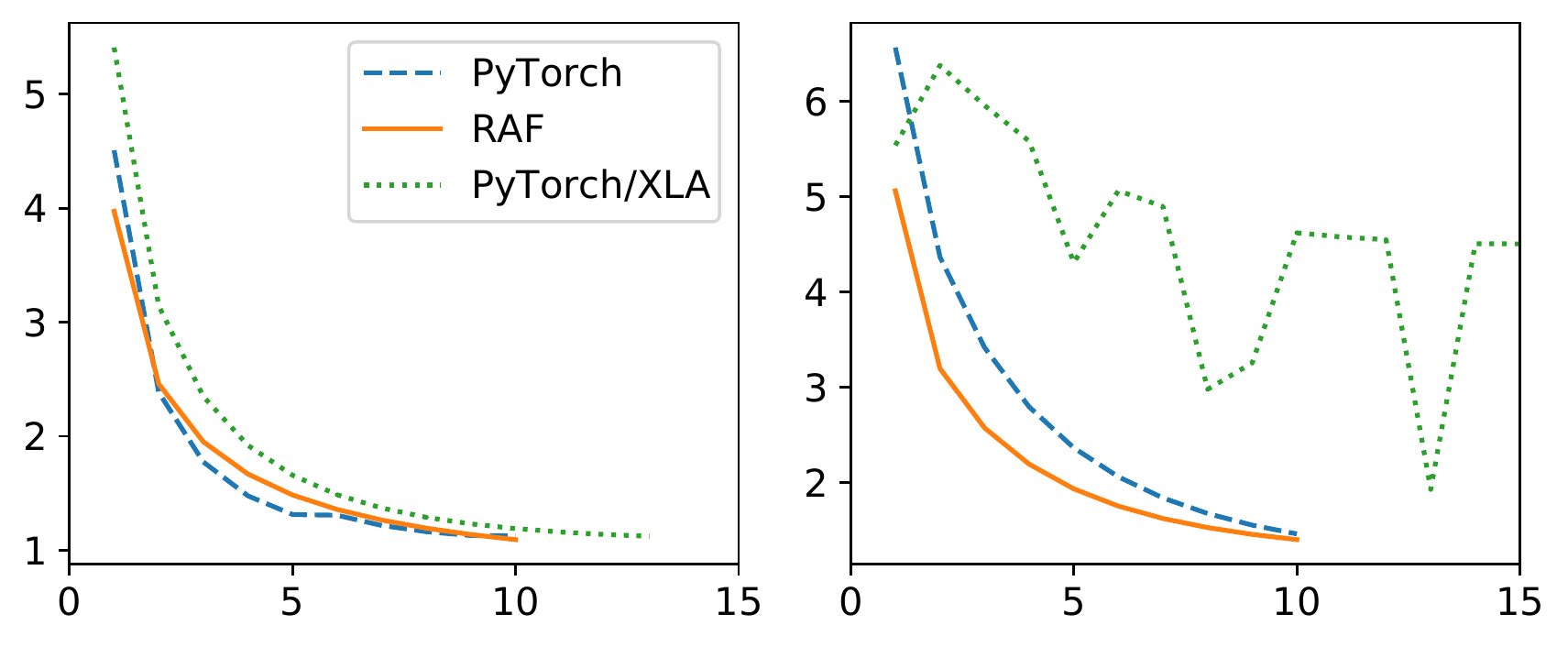}
    \caption{Loss (y-axis) vs. epochs (x-axis) when continue pre-training \textbf{Left:}BERT-large and \textbf{Right:}GPT-2 with AMP.}
    \label{fig:train}
\end{figure}

%% file: Tables/tbl-exp-train.tex
\begin{table}[!tbh]
\caption{Time-to-train for continued pre-training. Numbers are time-to-train (model setup time) in minutes.}
\centering
\begin{tabular}{c|cc}
\hline
Framework & BERT-Large & GPT-2 \\ \hline
RAF & 27.9 (6.2) & 11.9 (4.4) \\
PyTorch & 29.8 ($\sim$0) & 12.6 ($\sim$0) \\
PyTorch/XLA & 40 (5.3) & N/A (2.3) \\
\hline
\end{tabular}%
\label{tbl:exp-train}
\end{table}

%% file: Figures/fig-exp-dist.tex
\begin{figure}[tbh]
    \centering
    \includegraphics[width=0.47\textwidth]{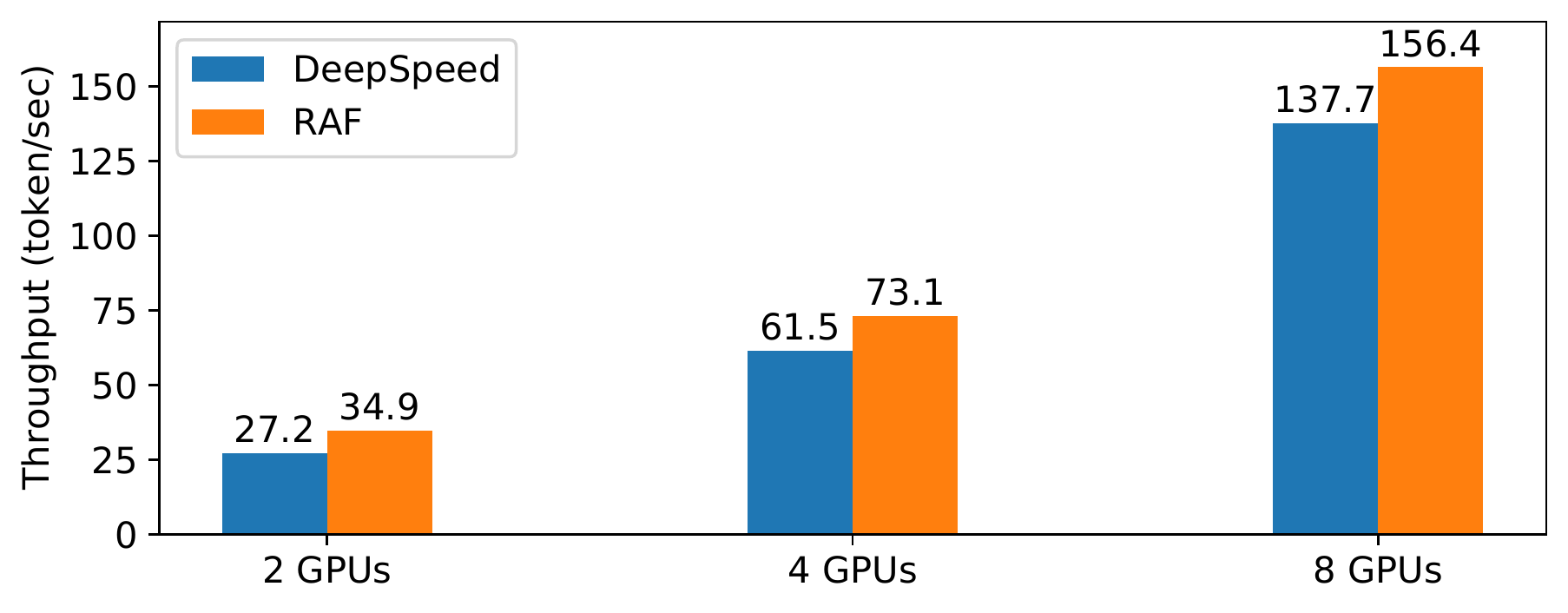}
    \caption{Throughput comparison when training a custom language model with 1.5B parameters on multiple GPUs.}
    \label{fig:dist}
\end{figure}

%% file: Figures/fig-exp-ablation-dist.tex
\begin{figure}[tbh]
    \centering
    \includegraphics[width=0.47\textwidth]{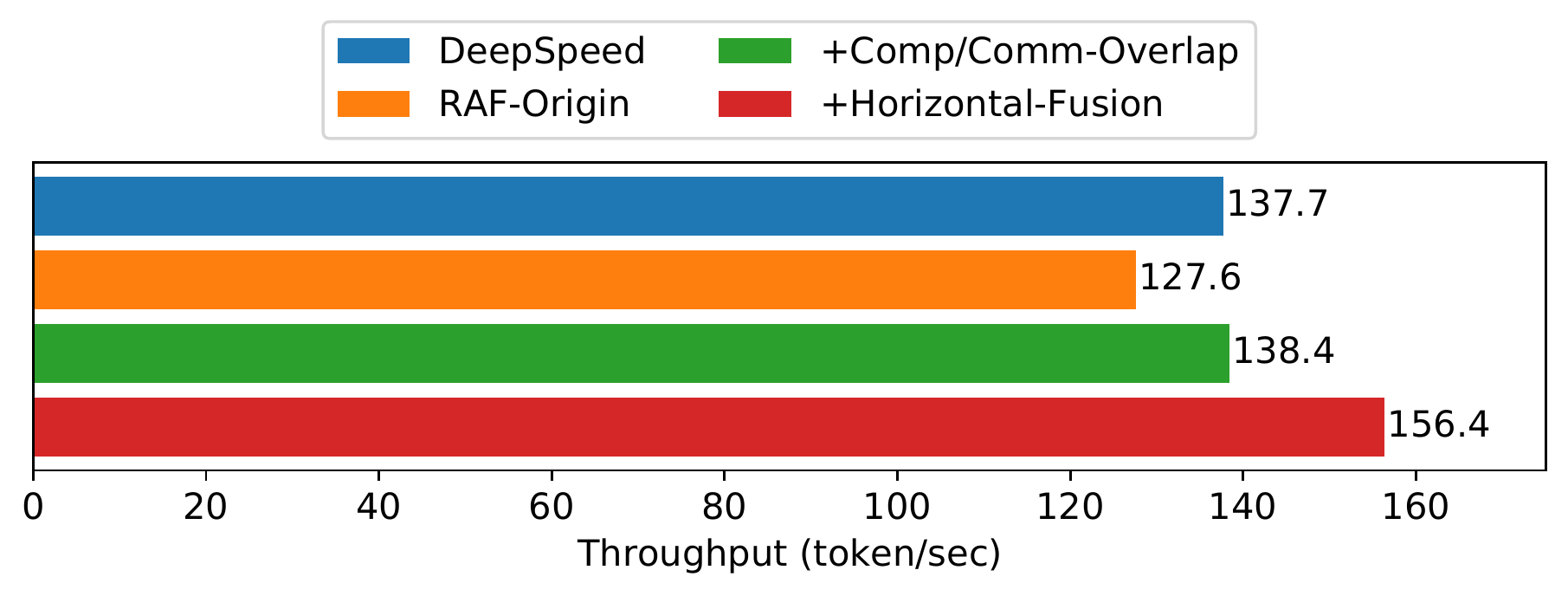}
    \caption{Ablation study of RAF to DeepSpeed on multiple GPUs.}
    \label{fig:ablation_dist}
\end{figure}

%% file: Figures/fig-exp-profile-dist.tex
\begin{figure}[tbh]
    \centering
    \includegraphics[width=0.47\textwidth]{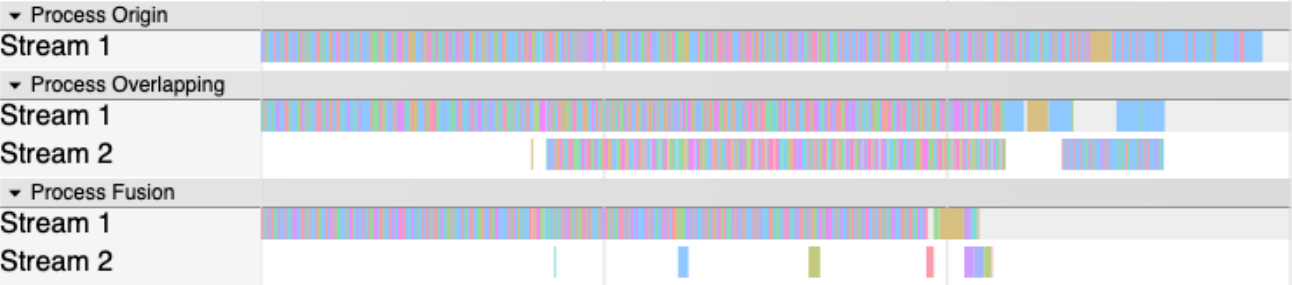}
    \caption{Operator-level profiling for RAF with custom model with 1.5 billion parameters on multiple GPUs.}
    \label{fig:dist_profile}
\end{figure}

%% file: Sections/6.related.tex
\section{Related Works}
\label{sec:related}
\vspace{-0.1cm}

\noindent\textbf{Deep Learning Compilers}
XLA~\cite{xla} is the most notable deep learning compiler in recent years. 
Deep learning framework such as TensorFlow~\cite{abadi2016tensorflow} and JAX~\cite{jax2018github} adopt XLA as their compiler backend.
Meanwhile, a recent work, LazyTensor~\cite{suhan2021lazytensor}, proposes lazy tensor core that bridges the gap between PyTorch and XLA. Although the focus in \cite{suhan2021lazytensor} is Google TPU, their open source framework, PyTorch/XLA, also supports the GPU backend.
When compare to \system, XLA relies on the training graph generated by the framework and hence sacrifices the graph level flexibility.
XLA adopts A-Normal Form (ANF) IRs for all optimizations, and dispatches operators to either kernel libraries (e.g., cuDNN~\cite{chetlur2014cudnn}) or code generation. Accordingly, it is tedious to implement certain optimizations, or plug in new emerging hand-crafted kernels.

In addition, there are several existing compilers that can be used to optimize deep learning workloads. 
Halide~\cite{ragan2013halide} proposes a domain-specific language and compiler originally for image processing pipeline. It is then applied to deep learning workloads~\cite{adams2019learning}.
TVM~\cite{chen2018tvm}, which is an open source deep learning compiler, shares many similarities with Halide but focuses more on deep learning and adopts a low-level representation, that explicitly expresses the choice of memory layout, parallelization pattern, locality and hardware primitives etc.
Tiramisu~\cite{baghdadi2019tiramisu}, is a polyhedral compiler for dense and sparse kernels and performs loop optimizations and data layout transformations with a scheduling language.
Astra~\cite{sivathanu2019astra} focuses the computation-intensive kernels like matrix multiplication; while leaving the other work such as element-wise operator fusion to the framework.

\noindent\textbf{Distributed Training}
There are several parallelism strategies when training a large model in a distributed fashion -- data parallelism and model parallelism.
For data parallelism, ZeRO~\cite{rajbhandari2020zero} is the most state-of-the-art work that reduces the replicated tensors of optimizer state, gradients and parameters. With ZeRO memory optimization, DeepSpeed~\cite{rasley2020deepspeed} successfully trains a transformer model with 100 billions of parameters. MiCS~\cite{zhang2022mics} further improves DeepSpeed by optimizing data communication for limited bandwidth on public clouds.
On the other hand, all above systems are based on transformer models and PyTorch, so they cannot be easily extended and generalized.
Meanwhile, ZeRO has been implemented as a compiler pass in \system, and our evaluation results have illustrated its efficiency against DeepSpeed.

For model parallelism, Megatron-LM~\cite{shoeybi2019megatron} manually partitions transformer models for distributed training.
GPipe~\cite{huang2019gpipe} is a pipeline parallelism library that partitions a network with a sequence of layers to multiple devices for training.
On the other hand, there are only a few existing work targeting distributed training compilation. Alpa~\cite{zheng2022alpa} is built on JAX~\cite{jax2018github} and XLA, to optimize inter- and intra-operator parallelism. Unlike \system that targets holistic optimizations, Alpa focuses on graph-level scheduling on distributed platforms and the idea can be integrated into \system in the future.

\noindent\textbf{Compiler Optimizations for Deep Learning}
While there are lots of compiler optimizations for deep learning workloads, we mainly discuss fusion and rematerialization due to their importance to training workloads.
The methods of \textit{operator fusion} can be classified into three categories: pattern-based, rule-based, and dependency-based fusion. The fixed pattern-based fusion restrains the combinations of operators that can be fused. It is adopted by NVIDIA CUTLASS~\cite{cutlass}, cuDNN~\cite{cudnnv8}, TensorRT~\cite{tensorrt}, and Intel oneDNN~\cite{onednn}.
However, patterns need to be defined ahead-of-time manually and fails to cope with new operators. 
In comparison, TVM~\cite{chen2018tvm}, XLA~\cite{xla}, DNNFusion~\cite{niu2021dnnfusion}, FusionStitching~\cite{zheng2020fusionstitching} embrace rule-based fusion that defines a set of rules over operators, helping expose more fusion.
However, one drawback of rule-based fusion is its flexibility to support multiple backends.
Lastly, dependency-based fusion leverages generic dependency analysis (e.g., polyhedral analysis~\cite{mullapudi2015polymage,bondhugula2008practical,verdoolaege2013polyhedral,zhao2022apollo}) to achieve the best generality, but faces scalability due to the NP-hard algorithms.
In \system, our fusion mechanism with operator dialect combines pattern-based and rule-based fusion, taking the advantages of both methods.

\textit{Rematerialization} reduces the peak memory of training a model by recomputing intermediate tensors in the backward pass. 
Chen et al.~\cite{chen2016training} propose a heuristic to train an $n$-layer linear network with $O(\sqrt{n})$ memory cost by partitioning the network into segments and only storing the outputs of each segment in the forward pass.
Gruslys et al.~\cite{gruslys2016memory} present a dynamic programming approach to select checkpoints for recurrent neural networks. 
Kumar et al.~\cite{kumar2019efficient} formulate rematerialization as an optimization problem and solves it with tree decomposition; while Checkmate~\cite{jain2020checkmate} leverages integer linear programming (ILP).
DTR~\cite{kirisame2020dynamic} is a runtime solution that maintains tensor metadata and uses carefully-designed cost functions to dynamically evict and rematerialize tensors during model execution.
On the other hand, DTR brings memory analysis overheads to the runtime, and may perform worse than static approaches on models without dynamism.
The approach \system adopted is similar to DTR but in compile time. In this way, we are able to conduct a complete liveness analysis to achieve smaller overheads than DTR.

%% file: Sections/7.conclusion.tex
\section{Conclusion}
\label{sec:concl}

This paper presents \system, a deep learning compiler for training. \system accepts vanilla models and performs in-house training graph generation, including automatic differentiation and mixed precision, to enable holistic optimizations for performance, memory and distributed training.
\system has an operator dialect mechanism that is capable of integrating third party kernel libraries as well as a tensor compiler, making sure to constantly catch up with the state-of-the-art kernel implementations.
The evaluation results show that \system achieves either better training throughput or larger batch sizes for popular transformer models on GPUs while preserving the numerically correctness.
The source code of \system is open source at \url{https://github.com/awslabs/raf}.

%% file: ms.bbl
\begin{thebibliography}{10}

\bibitem{abadi2016tensorflow}
Mart{\'\i}n Abadi, Paul Barham, Jianmin Chen, Zhifeng Chen, Andy Davis, Jeffrey
  Dean, Matthieu Devin, Sanjay Ghemawat, Geoffrey Irving, Michael Isard, et~al.
\newblock Tensorflow: A system for large-scale machine learning.
\newblock In {\em 12th {USENIX} Symposium on Operating Systems Design and
  Implementation OSDI 16)}, pages 265--283, 2016.

\bibitem{adams2019learning}
Andrew Adams, Karima Ma, Luke Anderson, Riyadh Baghdadi, Tzu-Mao Li,
  Micha{\"e}l Gharbi, Benoit Steiner, Steven Johnson, Kayvon Fatahalian,
  Fr{\'e}do Durand, et~al.
\newblock Learning to optimize halide with tree search and random programs.
\newblock {\em ACM Transactions on Graphics (TOG)}, 38(4):1--12, 2019.

\bibitem{baghdadi2019tiramisu}
Riyadh Baghdadi, Jessica Ray, Malek~Ben Romdhane, Emanuele Del~Sozzo,
  Abdurrahman Akkas, Yunming Zhang, Patricia Suriana, Shoaib Kamil, and Saman
  Amarasinghe.
\newblock Tiramisu: A polyhedral compiler for expressing fast and portable
  code.
\newblock In {\em 2019 IEEE/ACM International Symposium on Code Generation and
  Optimization (CGO)}, pages 193--205. IEEE, 2019.

\bibitem{bondhugula2008practical}
Uday Bondhugula, Albert Hartono, Jagannathan Ramanujam, and Ponnuswamy
  Sadayappan.
\newblock A practical automatic polyhedral parallelizer and locality optimizer.
\newblock In {\em Proceedings of the 29th ACM SIGPLAN Conference on Programming
  Language Design and Implementation}, pages 101--113, 2008.

\bibitem{bottou2010large}
L{\'e}on Bottou.
\newblock Large-scale machine learning with stochastic gradient descent.
\newblock In {\em Proceedings of COMPSTAT'2010}, pages 177--186. Springer,
  2010.

\bibitem{jax2018github}
James Bradbury, Roy Frostig, Peter Hawkins, Matthew~James Johnson, Chris Leary,
  Dougal Maclaurin, George Necula, Adam Paszke, Jake Vander{P}las, Skye
  Wanderman-{M}ilne, and Qiao Zhang.
\newblock {JAX}: Composable transformations of {P}ython+{N}um{P}y programs,
  2018.

\bibitem{chan2021speechstew}
William Chan, Daniel Park, Chris Lee, Yu~Zhang, Quoc Le, and Mohammad Norouzi.
\newblock Speechstew: Simply mix all available speech recognition data to train
  one large neural network.
\newblock {\em arXiv preprint arXiv:2104.02133}, 2021.

\bibitem{chen2018tvm}
Tianqi Chen, Thierry Moreau, Ziheng Jiang, Lianmin Zheng, Eddie Yan, Haichen
  Shen, Meghan Cowan, Leyuan Wang, Yuwei Hu, Luis Ceze, et~al.
\newblock {TVM}: An automated end-to-end optimizing compiler for deep learning.
\newblock In {\em 13th {USENIX} Symposium on Operating Systems Design and
  Implementation (OSDI 18)}, pages 578--594, 2018.

\bibitem{chen2016training}
Tianqi Chen, Bing Xu, Chiyuan Zhang, and Carlos Guestrin.
\newblock Training deep nets with sublinear memory cost.
\newblock {\em arXiv preprint arXiv:1604.06174}, 2016.

\bibitem{chetlur2014cudnn}
Sharan Chetlur, Cliff Woolley, Philippe Vandermersch, Jonathan Cohen, John
  Tran, Bryan Catanzaro, and Evan Shelhamer.
\newblock cu{DNN}: Efficient primitives for deep learning.
\newblock {\em arXiv preprint arXiv:1410.0759}, 2014.

\bibitem{chung2021w2v}
Yu-An Chung, Yu~Zhang, Wei Han, Chung-Cheng Chiu, James Qin, Ruoming Pang, and
  Yonghui Wu.
\newblock W2v-bert: Combining contrastive learning and masked language modeling
  for self-supervised speech pre-training.
\newblock In {\em 2021 IEEE Automatic Speech Recognition and Understanding
  Workshop (ASRU)}, pages 244--250. IEEE, 2021.

\bibitem{devlin2018bert}
Jacob Devlin, Ming-Wei Chang, Kenton Lee, and Kristina Toutanova.
\newblock Bert: Pre-training of deep bidirectional transformers for language
  understanding.
\newblock {\em arXiv preprint arXiv:1810.04805}, 2018.

\bibitem{gruslys2016memory}
Audrunas Gruslys, R{\'e}mi Munos, Ivo Danihelka, Marc Lanctot, and Alex Graves.
\newblock Memory-efficient backpropagation through time.
\newblock {\em Advances in Neural Information Processing Systems}, 29, 2016.

\bibitem{he2016deep}
Kaiming He, Xiangyu Zhang, Shaoqing Ren, and Jian Sun.
\newblock Deep residual learning for image recognition.
\newblock In {\em Proceedings of the IEEE conference on computer vision and
  pattern recognition}, pages 770--778, 2016.

\bibitem{huang2019gpipe}
Yanping Huang, Youlong Cheng, Ankur Bapna, Orhan Firat, Dehao Chen, Mia Chen,
  HyoukJoong Lee, Jiquan Ngiam, Quoc~V Le, Yonghui Wu, et~al.
\newblock Gpipe: Efficient training of giant neural networks using pipeline
  parallelism.
\newblock {\em Advances in neural information processing systems}, 32, 2019.

\bibitem{onednn}
Intel.
\newblock one{API} {D}eep {N}eural {N}etwork {L}ibrary (one{DNN}).
\newblock \url{https://github.com/oneapi-src/oneDNN}.
\newblock [Online; accessed 2022].

\bibitem{jain2020checkmate}
Paras Jain, Ajay Jain, Aniruddha Nrusimha, Amir Gholami, Pieter Abbeel, Joseph
  Gonzalez, Kurt Keutzer, and Ion Stoica.
\newblock Checkmate: Breaking the memory wall with optimal tensor
  rematerialization.
\newblock {\em Proceedings of Machine Learning and Systems}, 2:497--511, 2020.

\bibitem{jeon2021collage}
Byungsoo Jeon, Sunghyun Park, Peiyuan Liao, Sheng Xu, Tianqi Chen, and Zhihao
  Jia.
\newblock Collage: Automated integration of deep learning backends.
\newblock {\em arXiv preprint arXiv:2111.00655}, 2021.

\bibitem{kingma2014adam}
Diederik~P Kingma and Jimmy Ba.
\newblock Adam: A method for stochastic optimization.
\newblock {\em arXiv preprint arXiv:1412.6980}, 2014.

\bibitem{kirisame2020dynamic}
Marisa Kirisame, Steven Lyubomirsky, Altan Haan, Jennifer Brennan, Mike He,
  Jared Roesch, Tianqi Chen, and Zachary Tatlock.
\newblock Dynamic tensor rematerialization.
\newblock {\em arXiv preprint arXiv:2006.09616}, 2020.

\bibitem{krizhevsky2012alexnet}
Alex Krizhevsky, Ilya Sutskever, and Geoffrey~E Hinton.
\newblock Imagenet classification with deep convolutional neural networks.
\newblock In {\em Advances in neural information processing systems}, pages
  1097--1105, 2012.

\bibitem{kumar2019efficient}
Ravi Kumar, Manish Purohit, Zoya Svitkina, Erik Vee, and Joshua Wang.
\newblock Efficient rematerialization for deep networks.
\newblock {\em Advances in Neural Information Processing Systems}, 32, 2019.

\bibitem{lattner2021mlir}
Chris Lattner, Mehdi Amini, Uday Bondhugula, Albert Cohen, Andy Davis, Jacques
  Pienaar, River Riddle, Tatiana Shpeisman, Nicolas Vasilache, and Oleksandr
  Zinenko.
\newblock {MLIR}: Scaling compiler infrastructure for domain specific
  computation.
\newblock In {\em 2021 IEEE/ACM International Symposium on Code Generation and
  Optimization (CGO)}, pages 2--14. IEEE, 2021.

\bibitem{li2020pytorchdist}
Shen Li, Yanli Zhao, Rohan Varma, Omkar Salpekar, Pieter Noordhuis, Teng Li,
  Adam Paszke, Jeff Smith, Brian Vaughan, Pritam Damania, et~al.
\newblock Pytorch distributed: Experiences on accelerating data parallel
  training.
\newblock {\em arXiv preprint arXiv:2006.15704}, 2020.

\bibitem{liu2019roberta}
Yinhan Liu, Myle Ott, Naman Goyal, Jingfei Du, Mandar Joshi, Danqi Chen, Omer
  Levy, Mike Lewis, Luke Zettlemoyer, and Veselin Stoyanov.
\newblock Roberta: A robustly optimized bert pretraining approach.
\newblock {\em arXiv preprint arXiv:1907.11692}, 2019.

\bibitem{mattson2020mlperf}
Peter Mattson, Christine Cheng, Gregory Diamos, Cody Coleman, Paulius
  Micikevicius, David Patterson, Hanlin Tang, Gu-Yeon Wei, Peter Bailis, Victor
  Bittorf, et~al.
\newblock {MLP}erf training benchmark.
\newblock {\em Proceedings of Machine Learning and Systems}, 2:336--349, 2020.

\bibitem{merity2016pointer}
Stephen Merity, Caiming Xiong, James Bradbury, and Richard Socher.
\newblock Pointer sentinel mixture models.
\newblock {\em arXiv preprint arXiv:1609.07843}, 2016.

\bibitem{mullapudi2015polymage}
Ravi~Teja Mullapudi, Vinay Vasista, and Uday Bondhugula.
\newblock Polymage: Automatic optimization for image processing pipelines.
\newblock {\em ACM SIGARCH Computer Architecture News}, 43(1):429--443, 2015.

\bibitem{niu2021dnnfusion}
Wei Niu, Jiexiong Guan, Yanzhi Wang, Gagan Agrawal, and Bin Ren.
\newblock {DNNF}usion: Accelerating deep neural networks execution with
  advanced operator fusion.
\newblock In {\em Proceedings of the 42nd ACM SIGPLAN International Conference
  on Programming Language Design and Implementation}, pages 883--898, 2021.

\bibitem{amp}
Nvidia.
\newblock Automatic mixed precision for deep learning.
\newblock \url{https://developer.nvidia.com/automatic-mixed-precision}.
\newblock [Online; accessed 2022].

\bibitem{cudnnv8}
Nvidia.
\newblock cu{DNN} {R}elease 8.x.x.
\newblock
  \url{https://docs.nvidia.com/deeplearning/cudnn/release-notes/rel_8.html}.
\newblock [Online; 2020].

\bibitem{cutlass}
Nvidia.
\newblock {CUTLASS}: {CUDA} {T}emplates for {L}inear {A}lgebra {S}ubroutines.
\newblock \url{https://github.com/NVIDIA/cutlass}.
\newblock [Online; accessed 2017].

\bibitem{tensorrt}
Nvidia.
\newblock Nvidia {T}ensor{RT}.
\newblock \url{https://developer.nvidia.com/tensorrt}.
\newblock [Online; accessed 2022].

\bibitem{paszke2019pytorch}
Adam Paszke, Sam Gross, Francisco Massa, Adam Lerer, James Bradbury, Gregory
  Chanan, Trevor Killeen, Zeming Lin, Natalia Gimelshein, Luca Antiga, et~al.
\newblock Pytorch: An imperative style, high-performance deep learning library.
\newblock In {\em Advances in Neural Information Processing Systems}, pages
  8024--8035, 2019.

\bibitem{pearlmutter2008reverse}
Barak~A Pearlmutter and Jeffrey~Mark Siskind.
\newblock Reverse-mode {AD} in a functional framework: Lambda the ultimate
  backpropagator.
\newblock {\em ACM Transactions on Programming Languages and Systems (TOPLAS)},
  30(2):1--36, 2008.

\bibitem{nvfuser}
PyTorch.
\newblock
  \url{https://pytorch.org/blog/introducing-nvfuser-a-deep-learning-compiler-for-pytorch/}.
\newblock [Online; 2022].

\bibitem{radford2019language}
Alec Radford, Jeffrey Wu, Rewon Child, David Luan, Dario Amodei, and Ilya
  Sutskever.
\newblock Language models are unsupervised multitask learners.
\newblock {\em OpenAI blog}, 1(8):9, 2019.

\bibitem{ragan2013halide}
Jonathan Ragan-Kelley, Connelly Barnes, Andrew Adams, Sylvain Paris, Fr{\'e}do
  Durand, and Saman Amarasinghe.
\newblock Halide: A language and compiler for optimizing parallelism, locality,
  and recomputation in image processing pipelines.
\newblock {\em Acm Sigplan Notices}, 48(6):519--530, 2013.

\bibitem{rajbhandari2020zero}
Samyam Rajbhandari, Jeff Rasley, Olatunji Ruwase, and Yuxiong He.
\newblock Zero: Memory optimizations toward training trillion parameter models.
\newblock In {\em SC20: International Conference for High Performance
  Computing, Networking, Storage and Analysis}, pages 1--16. IEEE, 2020.

\bibitem{rasley2020deepspeed}
Jeff Rasley, Samyam Rajbhandari, Olatunji Ruwase, and Yuxiong He.
\newblock Deepspeed: System optimizations enable training deep learning models
  with over 100 billion parameters.
\newblock In {\em Proceedings of the 26th ACM SIGKDD International Conference
  on Knowledge Discovery \& Data Mining}, pages 3505--3506, 2020.

\bibitem{shoeybi2019megatron}
Mohammad Shoeybi, Mostofa Patwary, Raul Puri, Patrick LeGresley, Jared Casper,
  and Bryan Catanzaro.
\newblock Megatron-lm: Training multi-billion parameter language models using
  model parallelism.
\newblock {\em arXiv preprint arXiv:1909.08053}, 2019.

\bibitem{sivathanu2019astra}
Muthian Sivathanu, Tapan Chugh, Sanjay~S Singapuram, and Lidong Zhou.
\newblock Astra: Exploiting predictability to optimize deep learning.
\newblock In {\em Proceedings of the Twenty-Fourth International Conference on
  Architectural Support for Programming Languages and Operating Systems}, pages
  909--923, 2019.

\bibitem{suhan2021lazytensor}
Alex Suhan, Davide Libenzi, Ailing Zhang, Parker Schuh, Brennan Saeta,
  Jie~Young Sohn, and Denys Shabalin.
\newblock Lazytensor: Combining eager execution with domain-specific compilers.
\newblock {\em arXiv preprint arXiv:2102.13267}, 2021.

\bibitem{szegedy2015going}
Christian Szegedy, Wei Liu, Yangqing Jia, Pierre Sermanet, Scott Reed, Dragomir
  Anguelov, Dumitru Erhan, Vincent Vanhoucke, and Andrew Rabinovich.
\newblock Going deeper with convolutions.
\newblock In {\em Proceedings of the IEEE conference on computer vision and
  pattern recognition}, pages 1--9, 2015.

\bibitem{szegedy2016rethinking}
Christian Szegedy, Vincent Vanhoucke, Sergey Ioffe, Jon Shlens, and Zbigniew
  Wojna.
\newblock Rethinking the inception architecture for computer vision.
\newblock In {\em Proceedings of the IEEE Conference on Computer Vision and
  Pattern Recognition}, pages 2818--2826, 2016.

\bibitem{vaswani2017attention}
Ashish Vaswani, Noam Shazeer, Niki Parmar, Jakob Uszkoreit, Llion Jones,
  Aidan~N Gomez, {\L}ukasz Kaiser, and Illia Polosukhin.
\newblock Attention is all you need.
\newblock {\em Advances in neural information processing systems}, 30, 2017.

\bibitem{verdoolaege2013polyhedral}
Sven Verdoolaege, Juan Carlos~Juega, Albert Cohen, Jose Ignacio~Gomez,
  Christian Tenllado, and Francky Catthoor.
\newblock Polyhedral parallel code generation for cuda.
\newblock {\em ACM Transactions on Architecture and Code Optimization (TACO)},
  9(4):1--23, 2013.

\bibitem{wolf2019huggingface}
Thomas Wolf, Lysandre Debut, Victor Sanh, Julien Chaumond, Clement Delangue,
  Anthony Moi, Pierric Cistac, Tim Rault, R{\'e}mi Louf, Morgan Funtowicz,
  et~al.
\newblock Huggingface's transformers: State-of-the-art natural language
  processing.
\newblock {\em arXiv preprint arXiv:1910.03771}, 2019.

\bibitem{xla}
{XLA Team}.
\newblock Xla - tensorflow, compiled, March 2017.

\bibitem{zhang2020pushing}
Yu~Zhang, James Qin, Daniel~S Park, Wei Han, Chung-Cheng Chiu, Ruoming Pang,
  Quoc~V Le, and Yonghui Wu.
\newblock Pushing the limits of semi-supervised learning for automatic speech
  recognition.
\newblock {\em arXiv preprint arXiv:2010.10504}, 2020.

\bibitem{zhang2022mics}
Zhen Zhang, Shuai Zheng, Yida Wang, Justin Chiu, George Karypis, Trishul
  Chilimbi, Mu~Li, and Xin Jin.
\newblock {MiCS}: Near-linear scaling for training gigantic model on public
  cloud.
\newblock {\em arXiv preprint arXiv:2205.00119}, 2022.

\bibitem{zhao2022apollo}
Jie Zhao, Xiong Gao, Ruijie Xia, Zhaochuang Zhang, Deshi Chen, Lei Chen, Renwei
  Zhang, Zhen Geng, Bin Cheng, and Xuefeng Jin.
\newblock Apollo: Automatic partition-based operator fusion through layer by
  layer optimization.
\newblock {\em Proceedings of Machine Learning and Systems}, 4:1--19, 2022.

\bibitem{zheng2022alpa}
Lianmin Zheng, Zhuohan Li, Hao Zhang, Yonghao Zhuang, Zhifeng Chen, Yanping
  Huang, Yida Wang, Yuanzhong Xu, Danyang Zhuo, Joseph~E Gonzalez, et~al.
\newblock Alpa: Automating inter-and intra-operator parallelism for distributed
  deep learning.
\newblock {\em arXiv preprint arXiv:2201.12023}, 2022.

\bibitem{zheng2020fusionstitching}
Zhen Zheng, Pengzhan Zhao, Guoping Long, Feiwen Zhu, Kai Zhu, Wenyi Zhao,
  Lansong Diao, Jun Yang, and Wei Lin.
\newblock Fusionstitching: Boosting memory intensive computations for deep
  learning workloads.
\newblock {\em arXiv preprint arXiv:2009.10924}, 2020.

\end{thebibliography}
